\documentclass{article} % For LaTeX2e

\usepackage{hyperref}

\usepackage{amsmath}
\usepackage{amssymb}
\usepackage{mathtools}
\usepackage{amsthm}
\usepackage{url}
\usepackage{subcaption}
\usepackage{graphicx} 
\usepackage{float}
\usepackage{rotating}
\usepackage{array}

\usepackage[accepted]{icml2023}

\theoremstyle{plain}

\theoremstyle{definition}

\theoremstyle{remark}

\newcommand{\quotes}[1]{``#1''}
\renewcommand\algorithmicrequire{\textbf{Input:}}
\renewcommand\algorithmicensure{\textbf{Output:}}
\def\ve{{\boldsymbol{e}}}

\usepackage[textsize=tiny]{todonotes}

\icmltitlerunning{Decoding Layer Saliency in Language Transformers}

\begin{document}

\twocolumn[
\icmltitle{Decoding Layer Saliency in Language Transformers}



\icmlsetsymbol{equal}{*}

\begin{icmlauthorlist}
\icmlauthor{Elizabeth M.~Hou}{str}
\icmlauthor{Gregory Castanon}{str}
\end{icmlauthorlist}

\icmlaffiliation{str}{STR, 600 West Cummings Park, Woburn, MA 01801, USA}

\icmlcorrespondingauthor{Elizabeth M.~Hou}{elizabeth.hou@str.us}

\icmlkeywords{saliency, explainability, feature attribution, transformers, NLP}

\vskip 0.3in
]

\printAffiliationsAndNotice{ } 
\begin{abstract}
In this paper, we introduce a strategy for identifying textual saliency in large-scale language models applied to classification tasks.  In visual networks where saliency is more well-studied, saliency is naturally localized through the convolutional layers of the network; however, the same is not true in modern transformer-stack networks used to process natural language.  We adapt gradient-based saliency methods for these networks, propose a method for evaluating the degree of semantic coherence of each layer, and demonstrate consistent improvement over numerous other methods for textual saliency on multiple benchmark classification datasets. Our approach requires no additional training or access to labelled data, and is comparatively very computationally efficient.
\end{abstract}

\section{Introduction}
Trained on the vast swathes of open-source text available on the internet, large-scale language models have demonstrated impressive performance in text generation and classification.  Most recently, models with transformer-stack architectures have shown an impressive ability to focus on task-salient elements of language and utilize that focus to achieve superhuman performance in certain constrained areas. However, there is a growing concern that despite this performance, these models lack transparency and have unpredictable blind spots in certain areas. This has led to an increased focus on \textit{salience} in natural language i.e. identifying which elements of text the model considers important for making a decision.

Unlike computer vision, where the pixels relevant to a task are often grouped together, the words that are important in a movie review, article or resume may not be close to each other.  This lack of locality is reflected in the preferred architectures, with convolutional heads to visual networks encouraging local associations while stacks of fully connected transformer layers allow natural language tokens to associate more globally.  This free association, combined with the degree of model complexity in transformer architectures, leads to challenges in interpretability, as not all feature spaces within the hidden layers of the network map cleanly to natural language.  

Current methods that explain the decision making processes of transformer-stack architectures focus on the embedding layer.  However, these methods often result in confusing or redundant explanations, as information gets muddled passing through multiple layers of transformers in the stack. Along with \cite{rogers2020primer}, we hypothesize that a more meaningful, clear, decision-oriented representation exists in solely the later layers of the network. In this paper, we propose a method that only captures the signal of the later layers of the transformer stack and projects it back onto the token space of natural language. Our method can be paired with any layer-based saliency metric, explicitly accounts for multiple layers of self-attention mechanisms and reflects the implications of the complex pre-training and task specific fine-tuning on the layers in the architecture. We validate this method both with objective measures of model importance (Hiding / Revealing Game) and human measures of external consistency (Token Overlap), and demonstrate significant improvements over the state of the art.

Our contributions are:
\begin{enumerate}
    \item We propose a computationally-efficient method using a pre-trained language model (LM) head to \quotes{decode} the hidden layers by mapping their features back onto the token space and present its application to an example saliency approach (Grad-CAM \cite{selvaraju2017grad}) for both binary and multi-class classification problems.  Our method not only requires no additional training or labels and works for any saliency method that calculates layer-specific saliency, but prioritizes task-specific information over language structure.
    \item We demonstrate improvement over state of the art methods for natural language explainability in two objective measures.  In the \textit{Hiding / Revealing Game}, we show that the removal / addition of tokens we believe are important and damages / improves the performance of a network more than our competitors.  In \textit{Token Overlap}, we show that our method has dramatically fewer tokens that are important for multiple classes; indicating that our important tokens are truly indicative of the class in question.  Notably, we achieve these improvements despite not directly optimizing for either metric, suggesting a robust result.
\end{enumerate}

\section{Related Work}
The concept of feature attribution or saliency scores initially began in computer vision (CV) where the interest lay in being able to explain the object detection and classification decisions of convolutional neural networks (CNNs). A large swath of literature has arisen in the CV domain on model explainers that broadly fall into a several overlapping categories: gradient-based methods, propagation-based methods, and occlusion-based methods \cite{simonyan2014deep, zeiler2014visualizing, springenberg2014striving, bach2015pixel, noh2015learning, zhou2016learning, selvaraju2017grad, sundararajan2017axiomatic, shrikumar2017learning, lundberg2017unified, smilkov2017smoothgrad, fong2017interpretable, zhang2018top, gu2018understanding, omeiza2019smooth}; although some avoid these categories \cite{castro2009polynomial, ribeiro2016should}. 

The natural-language processing (NLP) community has adopted, extended, and introduced new variants of these methods for both simpler long short-term memory (LSTM) architectures \cite{li2016understanding, arras2017explaining, kadar2017representation} and more complex state-of-the-art transformer-stack-based architectures \cite{guan2019towards, Wallace2019AllenNLP, de2020decisions, chefer2021transformer, hase2021out, feldhus2021ThermoStat}. Additionally, with the language domain and their attention-based architectures came another category of explainability methods that either visualize or use attention weight values for explanations \cite{bahdanau2014neural, martins2016softmax, strobelt2018s, liu2018visual, thorne2019generating, kobayashi2020attention, hao2021self}. However, this new category did not come without its share of controversy, with many papers questioning or defending their explanatory power \cite{jain2019attention, wiegreffe2019attention, serrano2019attention, pruthi2019learning, vashishth2019attention, bastings2020elephant}. 

The transfer of explainability techniques from CNN and LSTM architectures to far larger and more complex architectures with stacks of multi-headed self-attention mechanisms has proven challenging. Unlike CNNs and LSTMs, the transformer stack has little cognitive motivation, instead relying on a pre-training regime over a massive corpora to learn language structure. Many works have attempted to derive meaning from the learned structures in the architecture including \cite{voita2019analyzing, michel2019sixteen} who study the role of multiple heads, other approaches (mentioned previously) study information flow in the self-attention mechanism as part of their attention based explainers, and still others simply probe and visualize the overall architecture \cite{tenney2019bert, kovaleva2019revealing, vig2019multiscale, jawahar2019does, rogers2020primer, likhosherstov2021expressive}. However, despite these works, explainability methods for transformer-stack architectures fail to account for a principal component of the architecture, the \textit{stacking of transformer blocks}. Unlike previous works, we propose a saliency method that explicitly accounts for its representation after multiple layers of self-attention mechanisms and reflects the implications of the complex pre-training and task specific fine-tuning on the layers in the architecture.

\section{Proposed Saliency Method}

We propose a method for producing explanations for decisions made by language models with encoder-based transformer architectures such as BERT \cite{devlin2018bert} and RoBERTa \cite{vaswani2017attention}. We extend saliency techniques from the CV domain to the NLP domain in a way that tracks the provenance of intermediate layers in the original input space. This allows our method to capture layer specific information about the contributions of each token in an input sequence to a language model's final decision.

In the following subsections, we use Grad-CAM as the driving example for a layer-wise saliency technique. However, as will be seen in Algorithm \ref{alg}, our method is agnostic to the specific saliency method as long as the method can calculate layer specific scores. We show that by using a language model head, we can compute interpretable explanations at any layer and for any saliency metric. The novelty of our overall method is the ability to assign explanatory power to the original tokens in the input sequence from scores calculated using information only downstream from a specific hidden layer in a transformer stack.

\subsection{Calculating Layer Saliency Scores}

One way to calculate scores that explain the contributions of a token is using Gradient-weighted Class Activation Mapping (Grad-CAM), a gradient-based saliency method which was first proposed in \cite{selvaraju2017grad} to produce visual explanations for CV problems. Grad-CAM is one of several gradient-based methods that has the advantage of being able to perform class-discriminative localization for any CNN-based models without requiring architectural changes or re-training.

We compute the gradient of the predicted score $y^c$ (before softmax) for class $c$ with respect to an output of a transformer block $h_l$. These gradients can be viewed as weights that capture the \quotes{importance} (for a specific class $c$) of each of the $K$ features for each element of the output sequence. Specifically for Grad-CAM, this represents a partial linearization of the model downstream from $h_l$. We calculate the Grad-CAM scores as a weighted combination of the features in the output and use a ReLU function to assign no importance to elements with negative scores. Explicitly, let 
\begin{flalign} 
& \alpha^c_l = \left[ \dots, \frac{\partial y^c}{\partial h^k_l} h^k_l, \dots \right] \,\,\quad \forall k= 1, \dots, K \label{feature_score} \\ 
& s^c_l = g(\alpha^c_l) = ReLU(\sum_{k=1}^K \alpha^{c,k}_l) \label{score}
\end{flalign}
where the $h^k_l$ are the column vectors corresponding to the $K$ features of the output of layer $l$ in the stack of transformer blocks, $\alpha^c_l$ is a matrix of size $n \times K$ whose elements are the scores for each of the features of the output $h_l$, and $s^c_l$ is vector of size $n$ whose elements are the scores aggregated over all features. 

Note, the scores $s^c_l$ can be replaced with any method\footnote{See Appendix (\ref{app}) for examples with other methods.} that calculates layer specific scores for each of the features by replacing \eqref{feature_score} with another matrix of layer specific feature scores and some other feature aggregation function $g(\cdot)$.

If we calculate the scores with respect to $0$-th layer outputs $h_0$ (embeddings of token, segment, and positional information), the elements of the score vector correspond directly to the tokens of the input sequence. However, once an input sequence passes through a transformer block (described explicitly in \eqref{selfattention} in Section \ref{ss:interpreting_hidden_layers}), this relationship no longer holds. Unlike CNNs where there is a clear provenance between the pixels and the outputs of the convolution filters, the multi-headed self-attention mechanism of transformers are far more complicated. The receptive fields of a CNN are local patches, whereas the receptive fields of the outputs of a transformer block are far more global consisting of the entire input. This is because each self-attention head uses the entire input to learn new representations for some subset of the features making each element of the output a function of all elements of the input. Many works \cite{vig2019multiscale, tsai2019transformer, likhosherstov2021expressive} have attempted to attribute meaning to the attention mechanism with varying levels of success; however, they primarily focus on a single transformer block. The meaning of an output sequence when the input sequence is passed through multiple transformer blocks in a stack is even less clear. 

Thus, while it is relatively easy to calculate scores $s^c_l$ with respect to the embeddings ($l = 0$), which already lie in the token space, it may not necessarily produce explanations that are most relevant to a models prediction. \cite{rogers2020primer} surveys 150 papers and derives potential explanations for the roles of the layers of the BERT model. They conclude that the lower layers have the most information about linear word order (i.e. the linear position of a word in a sentence \cite{lin-etal-2019-open}), the middle layers contain syntactic information, and the final layers are the most task-specific \cite{jawahar2019does}. Therefore, it would be worthwhile to also explore the explanatory power of the saliency scores of the other layers $l > 0$ where only information in the network downstream from that layer is included in the score. By only capturing information downstream from a specific layer, we ignore potentially task-irrelevant information in the earlier layers of the network. 

\subsection{Interpreting the Hidden Layers}
\label{ss:interpreting_hidden_layers}

In order to calculate saliency scores, e.g. \eqref{score}, that only capture information downstream from a specific layer, we need to project these scores into a space where the elements of the projected vector correspond directly to the original tokens of an input sequence. This allows us to have a meaningful version of the scores where we can directly understand the contributions of each token. 
And because the elements of a scores vector correspond directly to the elements of a transformer block's outputs, the problem of projecting scores into a token space is equivalent to the problem of projecting outputs into a token space. Thus this can be mathematically formulated as finding a mapping $f$ that minimizes the loss $\mathcal{L}(\cdot)$ between the output of a transformer block and its closest possible token space representation $t$,
\begin{flalign} \label{loss}
\arg\min_{f}{\mathcal{L}(t, f(h_l))}
\end{flalign}
where $f$ holds for any layer $l$ in a transformer stack and $t$ is a $n \times V$ right stochastic matrix whose rows lie in the token space $\mathcal{T}$ defined as the surface of a $V$-dimensional unit hypersphere. The axes of this hypersphere correspond to the $V$ tokens in a vocabulary, so any point on this hypersphere's surface is the weighted contributions of the tokens. 

One of the pre-training tasks of models with encoder-based transformer architectures (e.g. BERT) is the masked language model task, which is trained to minimize exactly this loss when $\mathcal{L}(\cdot)$ is a cross-entropy function, $t$ is the original input sequence, and the layer $l = L$ is the final transformer block in the stack. The masked language model task trains two functions: $f^{base}(\cdot)$ that represents the transformer-stack and $f^{lm}(\cdot)$ the language model (LM) head that minimizes $\mathcal{L}(t^i, f^{lm}(h^i_{L})) $ where $h^i_{L} = f^{base}(t^i) $ is the $i$-th element of the output of the full transformer stack and $t^i$ is a one-hot vector representing the token at the $i$-th element of the original input sequence.

While we now have a function $f^{lm}(\cdot)$ that solves a specific version of \eqref{loss}, we still need to understand the function's role in encoder-based transformer architectures. The LM head takes in a row $i$ of the $n \times K$ final transformer-stack output and transforms it to lie in the same space as the corresponding row of the $n \times V$ one-hot matrix of the original input sequence. It decomposes as
\begin{flalign} \label{weightedcombo}
f^{lm}(h^i_{L}) = \sum_{j=1}^V \hat{P}_{L}^{ij} \ve^{(j)}
\end{flalign}
where $\ve^{(j)}$ is a $1 \times V$ basis vector with a one in column $j$ and a zero elsewhere representing $j$-th dimension of the token space $\mathcal{T}$ and $\hat{P}_{L}$ is a $n \times V$ right stochastic matrix with each row $i$ containing the (after softmax) prediction probabilities of being the $j$-th token in the vocabulary. Thus, because the basis vector $\ve^{(j)}$s correspond to tokens where $j$ is the token's position in the vocabulary, we can interpret $\hat{P}_{L}^{ij}$ as the amount of influence the $j$-th token has on $i$-th element of $h_{L}$. 

However, the masked language model task is training a function specifically for the final output of the transformer stack $h_{L}$. In order to extrapolate the effects of the $f^{lm}(\cdot)$ function to the output of any layer $l$, we must understand the most complicated part of an encoder-based transformer architecture, the self-attention mechanism. As previously studied in \cite{likhosherstov2021expressive, tsai2019transformer}, the output of the self-attention mechanism can be expressed as
\begin{flalign} \label{selfattention}
X' = A X W_V
\end{flalign}
where $A=softmax ( \frac{X W_Q W_K^T X^T}{\sqrt{d}} )$ is the normalized self-attention matrix, $W_Q, W_K, W_V$ are the query, key, and value weight matrices, and $d$ is hidden dimension of the self-attention mechanism. The self-attention matrix $A$ is a weighted similarity or kernel gram matrix between the elements of the input $X$, and the features of the output $X'$ are weighted combinations of the features of the inputs. The multi-headed mechanism simply combines various self-attention mechanisms in a weighted fashion. The rest of the transformer block consists of a feed-forward component and some layer additions and normalizations. Thus we can describe outputs of a transformer block overall as approximately a weighted combination of its inputs. Similarly, stacking transformer blocks together results in further weighted combinations of the original input sequence. 

\begin{figure}[H]
\begin{center}
\includegraphics[width=\linewidth]{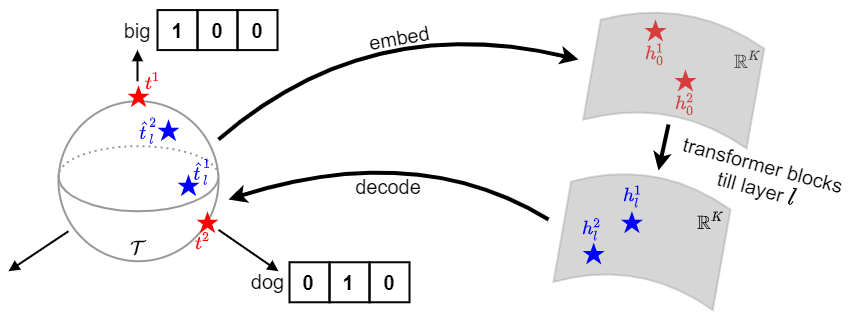}
\end{center}
\caption{The input sequence \quotes{big dog} is tokenized into two tokens $t^1$ and $t^2$ that lie on the unit hypersphere $\mathcal{T}$ and are then embedded into having $K$ continuous features ($h^1_0$ and $h^2_0$). A series of $l$ transformer blocks is applied to the embedded input sequence to produce $h^1_{l}$ and $h^2_{l}$, which are decoded back onto $\mathcal{T}$ with the $f^{lm}(\cdot)$ function. The outputs ($\hat{t}^1_l$ and $\hat{t}^2_l$) of the transformer blocks in $\mathcal{T}$ can be interpreted as weighted combinations of the original tokens $t^1$ and $t^2$.}
\label{fig:map}
\end{figure}

This leads to the key idea that because the outputs of any stack of transformer blocks are a weighted combination of original $K$ features, they lie in the same continuous feature space $\mathbb{R}^{K}$. Unlike the original token space $\mathcal{T}$, this feature space does not have an easily interpretable meaning. We conjecture that because the pre-training tasks are performed over an enormous corpus (Wikipedia etc.), the learned function $f^{lm}(\cdot)$ is estimating the map between $\mathbb{R}^{K}$ and $\mathcal{T}$ where the feature space is much smaller than the token vocabulary space $K << V$. Thus the $f^{lm}(\cdot)$ function acts as a universal decoder that predicts the likeliest combination of tokens, i.e. basis vectors $\ve^{(j)}$ of $\mathcal{T}$, that make up $f^{lm}(h_{l})$ where $\hat{P}_{l}$ are the prediction probabilities. We provide a simple example illustrating this process in Figure \ref{fig:map} to provide geometric intuition.

While the rows of $\hat{P}_{l}$ can be interpreted as the likelihood that each element of $h_l$ is a certain token in the vocabulary, the columns of $\hat{P}_{l}$ can analogously be interpreted as the amount of influence each token has on $h_l$. However, because we are only interested in the contributions of tokens from the input sequence, we can subset the $V$ columns of $\hat{P}_{l}$ to only the $T$ columns that correspond to the unique input sequence tokens. Let $\hat{D}_{l}$ be a $T \times n$ matrix where the rows of $\hat{D}_{l}$ are the columns of $\hat{P}_{l}$ that correspond to the tokens in the input sequence and the columns indices of $\hat{D}_{l}$ correspond to the elements of $h_l$. Now that we have a way to account for the contributions of tokens in an input sequence to a hidden layer, we can calculate a layer's saliency scores with respect to these tokens as
\begin{flalign} \label{proj_scores}
s^c_l = ReLU(\sum_{k=1}^K \hat{D}_{l} \alpha^{c,k}_l) 
\end{flalign}
where the elements of $s^c_l$ are weighted combinations of the features scores for layer $l$ from \eqref{feature_score} with each element being a different weight according to the rows of $\hat{D}_{l}$. This method can also be viewed as a generalization of \eqref{score} where the feature aggregation function $g(\cdot)$ has been modified to include additional weights.

These saliency scores $s^c_l$ capture the importance of each token in the input sequence to the models decision using \textit{only} information in the model that is downstream from a specific layer $l$ in a transformer stack. Thus we can view the layer choice $l$ as a control for the amount of model information used in a saliency score.

Additionally if we only want to allow contributions from the most important tokens, we can restrict each $\hat{s}^i_l$ in \eqref{proj_scores} to be a weighted combination of only the output scores where the input token $t_i$ is an top ranked contributor. We show pseudocode for our approach in Algorithm \ref{alg} which takes as inputs a tokenized input sequence $t$, a layer choice $l$, a threshold $\tau$ for the number of contributions from top ranked tokens, a feature aggregation function for a saliency method $g(\cdot)$, a LM task head from a pre-trained model $f^{lm}(\cdot)$, and a fine-tuned classification model $m(\cdot)$ where $m_l^{base}(\cdot)$ is the output of layer $l$ in the transformer stack and $m^{class}(\cdot)$ is the classification task head. For a given layer $l$, estimate the features scores $\hat{\alpha}_l$ for the most likely prediction $\hat{y}$ and the output of the $l$-th block in the transformer stack $h_l$ (Lines 1-3). Then estimate the token contributions $\hat{D}_{l}$ as a subset of $\hat{P}_{l}$, the LM probability predictions for $h_l$ onto the input token sequence $t$ (Lines 4-5). Finally the saliency scores $\hat{s}_l$ are a weighted sum of output scores $\hat{\alpha}_l$ where the non-zero weights are the top $\tau$ ranked values in the columns of $\hat{D}_{l}$ (Lines 6-12). These scores are then aggregated with a feature aggregation function $g(\cdot)$, for Grad-CAM $g(\cdot) = ReLU(\sum_{k=1}^K \cdot)$.

\begin{algorithm}[H]
\caption{Transformer-stack architectures embed a discrete vocabulary into a lower dimensional continuous space where layers in the stack merely transform it within this space. Our approach generates a decoder from the low-dim latent space back to the original token space.}
\label{alg}
\begin{algorithmic}[1]
  \REQUIRE $t, l, \tau, g(\cdot), f^{mlm}(\cdot), m(\cdot)$
  \STATE $h_{l} = m_l^{base}(t)$
  \STATE $\hat{y} = \max{m^{class}(m^{base}(t))}$
    \STATE $\hat{\alpha}_l = \frac{\partial \hat{y}}{\partial h^k_l} h^k_l$ or some other feature score for layer $l$
    \STATE $\hat{P}_{l} = f^{lm}(h_{l}) $
    \STATE $\hat{D}^{i}_{l} = (\hat{P}^j_{l})^\top \,\quad \forall j$ corresponding to tokens $t^i$
    \FOR{$i = 1$ to $T$}
        \FOR{$j = 1$ to $n$}
            \IF{$\hat{D}^{ij}_{l}$ in top $\tau$ ranked values of $\hat{D}^{j}_{l}$}
                \STATE $\hat{s}^j_l \mathrel{+}= g(\hat{D}^{ij}_{l} \hat{\alpha}^j_l)$ 
            \ENDIF
        \ENDFOR
    \ENDFOR
\ENSURE $\hat{s}_l$
\end{algorithmic}
\end{algorithm}

While the weights for the classification model $m(\cdot)$ change when fine-tuned to a specific dataset, they are still initialized at the pre-trained values. In practice because the weights of the model base $m_l^{base}$ change very little during fine-tuning, and in some training schemes are restricted not to change, we can still assume the transformer block outputs $h_{l}$ will still lie in $\mathbf{R}^{K}$. Thus because $f^{lm}(\cdot)$ estimates the map from $\mathbf{R}^{K}$ to $\mathcal{T}$, it is still able to decode $h_{l}$ despite being the outputs of a model with different weights. 

\section{Experiments}

In this section, we detail our experimental results on two benchmark classification task datasets\footnote{For a full description of the datasets, see Section \ref{exp_setup}.}: SST-2 \cite{socher2013recursive} a binary classification dataset that is one of the the General Language Understanding Evaluation (GLUE) \cite{wang2019glue} tasks and AG News \cite{zhang2015character} a subset (4 largest classes) of news articles from more than 2,000 news sources gathered by \cite{agnews}. We implemented our approach (labelled Decoded Grad-CAM) on a RoBERTa base from HuggingFace \cite{wolf2020transformers} using Grad-CAM for the saliency scores\footnote{See Appendix (\ref{game_other_results}) for additional examples using the  method from \cite{simonyan2014deep} for saliency scores.} and compared against numerous other explainability methods that have been  trained and provided by the AllenNLP Interpret \cite{Wallace2019AllenNLP} and ThermoStat \cite{feldhus2021ThermoStat} Python packages. To the best of our abilities, we have attempted to mimic the training regimes described by their respective packages for all competing models' explainability methods. However, in order to maintain consistency across all experiments and improve visibility, we have chosen to always use the standard RoBERTa base model with a 12 layer transformer stack. For further details on the experimental setup see Section \ref{exp_setup}.

\subsection{The Hiding /Revealing Game}

In order to evaluate the explainability of a token, we use the Hiding Game \cite{fong2017interpretable, castanon2018visualizing} and an inverse variant of it, which we will call the Revealing Game. For NLP, the Hiding Game iteratively obscures the least important tokens according to some score attributed with the token, replaces them with a [MASK] token, and removes them from the self-attention mechanism. The Revealing Game does the opposite and starts with a completely masked sequence and iteratively reveals the most important tokens according to their score. For both games, the prediction accuracy is periodically calculated at percentages of the total sequence length (ignoring [PAD] tokens). Similar variants such as positive / negative perturbations \cite{chefer2021generic} or using masking in \cite{hase2021out} have also been used for evaluating the explainability of methods. In addition to the AllenNLP Interpret and ThermoStat explainers, we also compare against a random baseline that is averaged over 20 random perturbations.

\begin{figure}[h]
\centering
\begin{subfigure}{\linewidth}
    \centering
    \includegraphics[width=\linewidth]{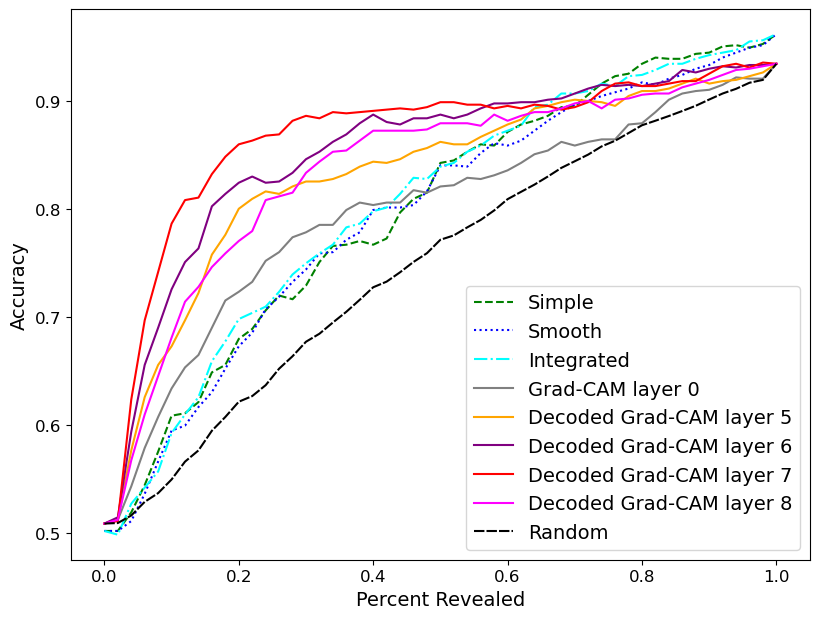}
    \caption{Revealing Game}
\end{subfigure}

\vspace{16pt}

\begin{subfigure}{\linewidth}
    \centering
    \includegraphics[width=\linewidth]{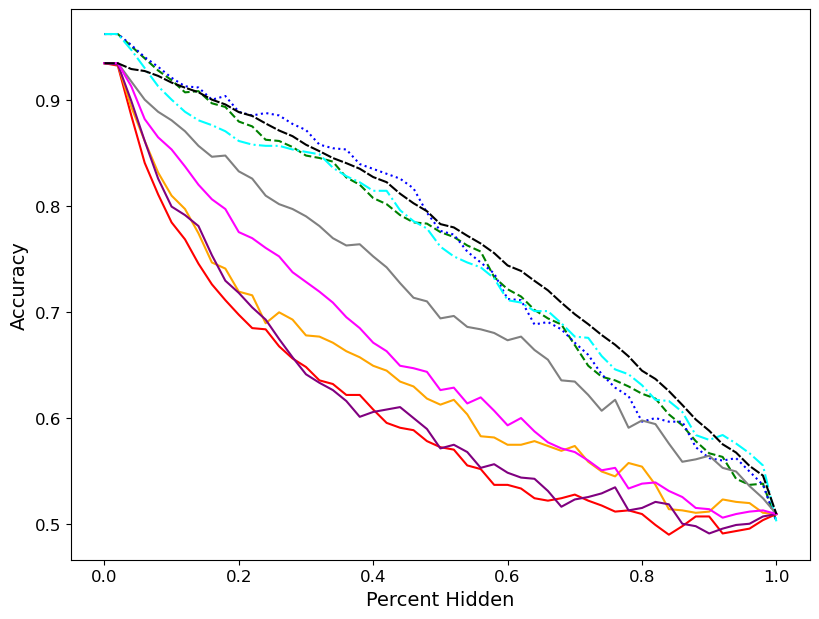}
    \caption{Hiding Game}
\end{subfigure} 
\caption{Our Decoded Grad-CAM method against vanilla Grad-CAM, AllenNLP Interpret explainers, and a random baseline on the SST-2 binary sentiment classification dataset.}
\label{sst_hiding_game}
\end{figure}

In Figure~\ref{sst_hiding_game}, we use the Hiding / Revealing Game to evaluate the accuracy of various explainability methods on the SST-2 sentiment classification dataset. We show the performance of the best layers\footnote{See Section \ref{all_layers} for all layers} of our Decoded Grad-CAM (for visibility) against AllenNLP Interprets implementation of the Simple, Smooth, and Integrated methods along with a layer 0 (vanilla) version of Grad-CAM.

For the Revealing Game, the accuracy of our layers shoots up very quickly after the first couple of tokens are revealed, implying that those tokens are very important to the model's classification decision. For the Hiding Game, all four of our layers have steeper drops in accuracy, which implies that the tokens being masked are more important as they dramatically affect the accuracy. Note that we are using a RoBERTa-base model, which is smaller than the RoBERTa-large model used by the AllenNLP Interprets explainers and is the reason for the small gap in accuracy for the full input sequence. Despite having a smaller underlying model, our Decoded Grad-CAM at layer 7 outperforms the explainers on a larger model up until the vast majority ($\approx$ 70\%) of important tokens have been revealed. 

\begin{figure}[H]
\centering
\begin{subfigure}{\linewidth}
    \centering
    \includegraphics[width=\linewidth]{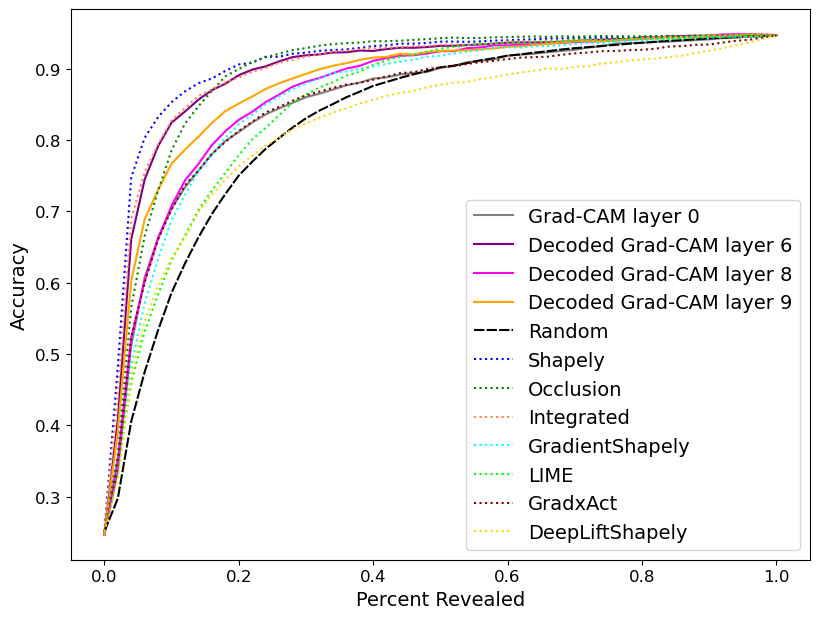}
    \caption{Revealing Game}
\end{subfigure}

\vspace{16pt}

\begin{subfigure}{\linewidth}
    \centering
    \includegraphics[width=\linewidth]{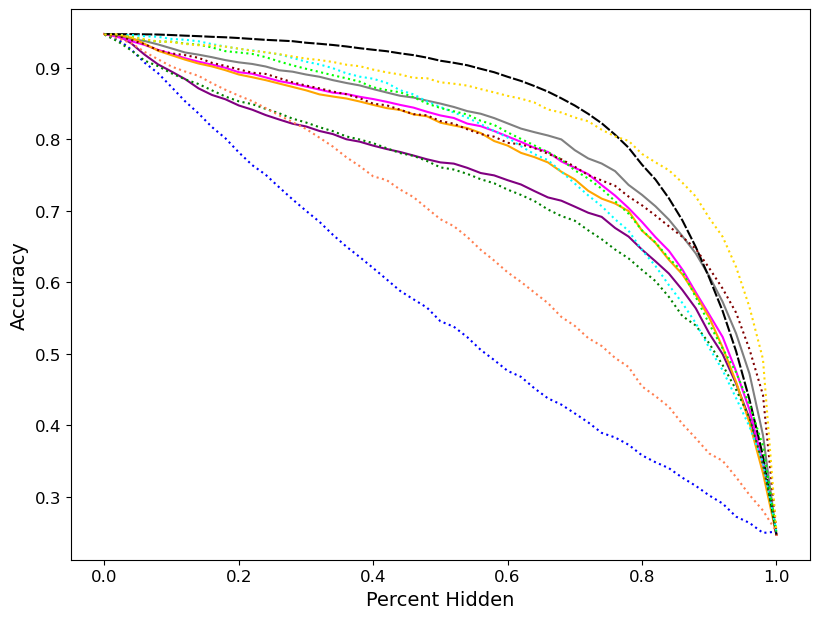}
    \caption{Hiding Game}
\end{subfigure} 
\caption{Our Decoded Grad-CAM method against vanilla Grad-CAM, ThermoStat explainers, and a random baseline on the AG News four topic classification dataset.}
\label{ag_hiding_game}
\end{figure}

We also apply the Hiding / Revealing Game to the AG News dataset, which is a multi-class topic classification task, and evaluate against numerous ThermoStat explainers. We show the accuracy of the best layers\footnote{See Section \ref{all_layers} for all layers} against all the ThermoStat explainers in Figure~\ref{ag_hiding_game}.

Many of the ThermoStat explainers are specifically built to probe for changes in predictions from changes in tokens and are essentially optimized to do well in the Hiding Game. However due to this perturbation construction, many of these methods (Integrated, LIME, Occlusion, and Shapely) are also extremely computationally expensive requiring many passes forward through the model and the ThermoStat package was specifically constructed to improve accessibility (at least for benchmark datasets) to these explainers \cite{feldhus2021ThermoStat}. In contrast, our decoded layer saliency method is applicable to any trained model without requiring any re-training and only requires one backward pass, which is far more computationally efficient. Additionally, because our method is not probing for changes from a correct class to an incorrect one, it does not require labels. This makes it useful as an explainer even in scenarios where a user only has access to a trained model and does not have access to any training or labelled data.

Despite this, our Decoded Grad-CAM layer 6 outperforms the majority of the ThermoStat explainers including the computationally heavy LIME method. Similar to the results for the SST-2 dataset, we see the accuracy of our best layers shoot up quickly in the Revealing Game with layer 6 having very competitive results with ThermoStats' Shapely, Occlusion, and Integrated methods. For the Hiding Game, our layers do not exhibit as dramatic of a drop, but still do significantly better than most ThermoStat explainers and our layer 6 is competitive with the all except the Shapely method until $\approx$ 40\% of tokens are hidden.

We also calculate the Area Under the Curve (AUC) for the explainers in both figures above in Table~\ref{auc}, where the best layer explainer of our method is \textbf{bolded} and the best competing explainer is \textbf{\textit{italicized}}. From Figure~\ref{ag_hiding_game}'s AUCs, for the Revealing Game, the difference in performance between the top four best explainers is extremely minor with only 0.011 gap between the first and fourth place methods. For the Hiding Game, the Shapely explainer is clearly the best; however our layer 6 still has respectable performance being only 0.008 worse than the Occlusion method and 0.078 worse than the Integrated method. 

Another noteworthy observation is that our best performing layers (5-8 for the SST-2 dataset and 6, 8 and 9 for the AG News dataset) roughly correspond to the \quotes{middle layers} described by \cite{rogers2020primer}. Thus our saliency method is only including information downstream from these layers, namely those corresponding to the \quotes{later layers}, which are described to be more task-specific. These conclusion are also supported by experiments in \cite{jawahar2019does} where they conclude that the \quotes{later layers} of the model are capture information about semantic tasks. Semantic tasks (e.g. randomly replacing nouns/verbs or swapping clausal conjuncts) often operate on parts of text that are also the most useful for classification; thus we are in agreeance about the role of these \quotes{later layers}.

\begin{table}[h]
\caption{\underline{Area Under the Curve for Revealing Game ($\uparrow$ is better)} \\ \underline{and Hiding Game ($\downarrow$ is better)}}
\label{auc}
\vskip\baselineskip
\begin{minipage}{\linewidth}
\centering
\begin{tabular}{l|c|c}
Explainer & Revealing & Hiding \\ \hline 
& & \\
Grad-CAM $l_0$ & 0.797 & \textit{\textbf{0.713}} \\
Decoded Grad-CAM $l_5$ & 0.832 & 0.64 \\
Decoded Grad-CAM $l_6$ & 0.854 & 0.617 \\
Decoded Grad-CAM $l_7$ & \textbf{0.867} & \textbf{0.609} \\
Decoded Grad-CAM $l_8$ & 0.836 & 0.66 \\
Simple & 0.799 & 0.756 \\
Smooth & 0.795 & 0.762 \\
Integrated & \textit{\textbf{0.804}} & 0.756 \\
Random & 0.748 & 0.77 \\
\end{tabular}
\subcaption{SST-2 Dataset (Figure~\ref{sst_hiding_game})}
\label{auc_sst}
\end{minipage}
\hfill
\vskip\baselineskip
\begin{minipage}{\linewidth}
\centering
\begin{tabular}{l|c|c}
Explainer & Revealing & Hiding \\ \hline 
& & \\
Grad-CAM $l_0$ & 0.853 & 0.804 \\
Decoded Grad-CAM $l_6$ & \textbf{0.895} & \textbf{0.739} \\
Decoded Grad-CAM $l_8$ & 0.867 & 0.782 \\
Decoded Grad-CAM $l_9$ & 0.879 & 0.774 \\
Shapely & \textbf{\textit{0.905}} & \textit{\textbf{0.567}} \\
Occlusion & 0.894 & 0.731 \\
Integrated & 0.897 & 0.661  \\
GradientShapely & 0.86 & 0.784 \\
LIME & 0.851 & 0.792 \\
GradxAct & 0.851 & 0.794 \\
DeepLiftShapely & 0.823 & 0.842 \\
Random & 0.829 & 0.841 \\
\end{tabular}
\subcaption{AG News Dataset (Figure~\ref{ag_hiding_game})}
\label{auc_ag}
\end{minipage}
\end{table}

\subsection{Token Overlap}

While the previous experiments are a good way to evaluate the affect of tokens on a model's decision making, they don't actually provide any indication of explainability to a human. In order to judge the human intuitiveness of the explanations, we should also consider the actual meanings of the top ranking tokens. Thus, we aggregate the scores of all tokens for all input sequences in a predicted class and weight their total score by how rarely they occur in everyday language i.e. the inverse document frequency of a random collection of 50,000 Wikipedia articles. The intuition behind this is that tokens that have high importance scores \textit{and} occur often in the input sequence of a predicted class relative to usage in common language are representative of that class. By aggregating over all input sequences of a predicted class, we also reduce the rewarding of one-off tokens that only explain the model's decisions for that specific input sequence. We can visualize the most important tokens for each predicted class in word clouds shown in Figures \ref{fig:sst2_word_clouds} and \ref{fig:ag_news_word_clouds} in Section \ref{word_clouds}.

Additionally, for classification tasks, tokens should disambiguate classes. So tokens that are important to a predicted class should be \textit{indicative}, i.e. unique to a class, especially if the classes are complements of each other. For example, tokens that are strong indicators that a movie review is positive should not also be strong indicators that a movie review is negative. Therefore, we can also evaluate the representativeness of tokens deemed important to a predicted class by considering how often they appear in multiple classes. Explicitly, we count the number of top $k$ ranked tokens that appear in every pair of classes and divide by the total count in order to get the percentage of token overlap. 

\begin{figure}[H]
\centering
\begin{subfigure}{\linewidth}
    \centering
    \includegraphics[width=\linewidth]{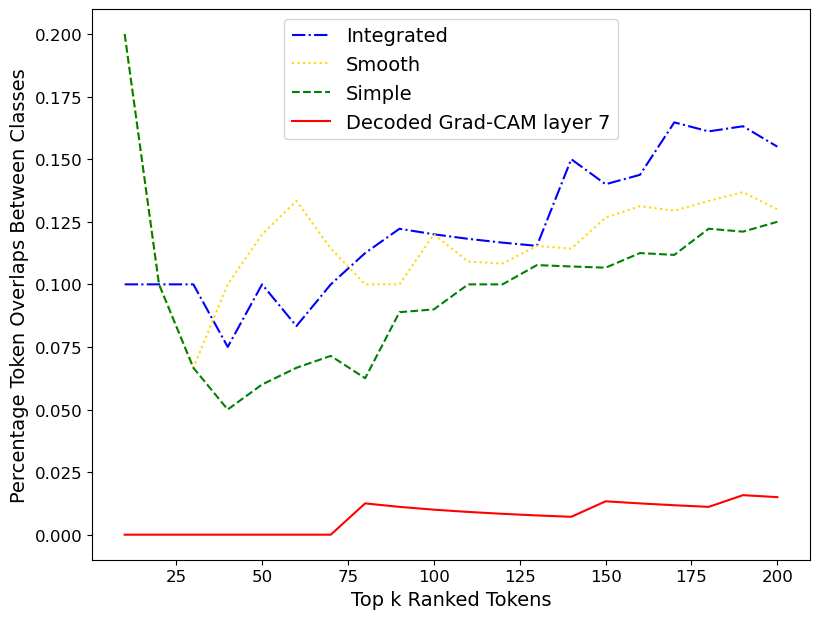}
\caption{SST-2 dataset}
\end{subfigure}

\vspace{16pt}

\begin{subfigure}{\linewidth}
    \centering
    \includegraphics[width=\linewidth]{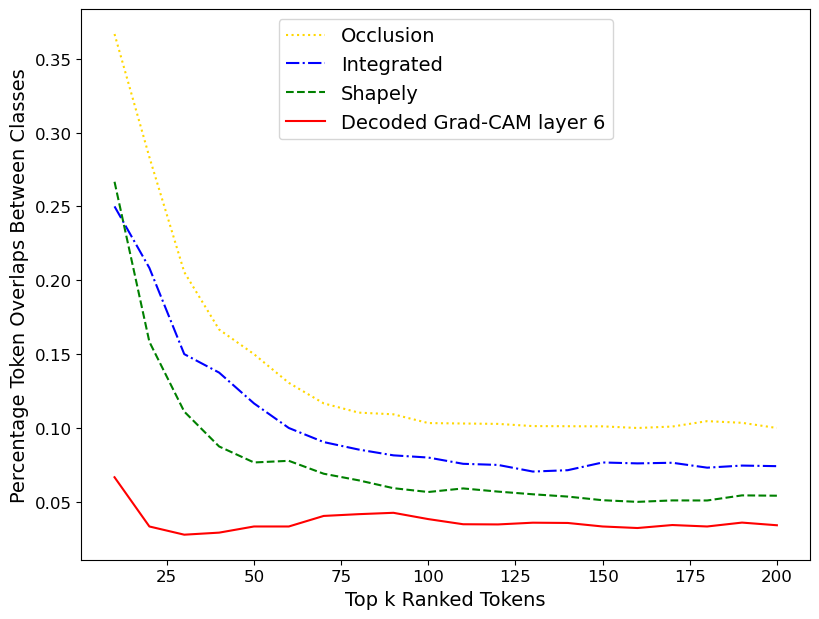}
\caption{AG News dataset}
\end{subfigure} 
\caption{Percentage of tokens that appear in multiple classes for the top $k$ most important tokens.}
\label{fig:tok_k}
\end{figure}

We show this percentage as a function of the top $k$ ranked tokens in Figure~\ref{fig:tok_k} for the best Decoded Grad-CAM layer according the Hiding / Revealing Game against the AllenNLP Interpret explainers on the SST-2 dataset and against the best ThermoStat explainers on the AG News dataset. Additionally we show, in tables in Section \ref{overlap}, the actual tokens in the top 50 that appear in multiple classes of the SST-2 and AG News datasets respectively, along with the raw counts of token overlap in Figure~\ref{fig:tok_count_k}.

For both datasets, our best Decoded Grad-CAM layer significantly outperforms the competing methods with very few important tokens belonging to multiple classes. Unlike the other explainability methods, our approach only incorporates information in the network that is downstream from a specific layer. Thus its does not include language structure information such as the word order or syntactic information from earlier layers that would add noise to the explainer. We can interpret from the plots that because the competing methods have many more tokens that are salient for multiple classes, these tokens may be structurally important, but not class discriminate. The removal of these structurally important tokens may also be causing an out of distribution effect in the Hiding Game \cite{hase2021out} and biasing their good performance. We also provide some example snippets of input sequences highlighted by the methods in the above figures in tables in Section \ref{highlights}. These examples provide an additional way for a human to directly visualize and interpret the explainability of the methods for a particular input sequence.

\section{Discussion}

In this paper, we have presented an approach for measuring the importance of tokens to a classification task based on the information encoded in the hidden layers of the transformer stack. Consistent with previous research into the meanings of these intermediate layers in large-scale language models, we explicitly confirm, through multiple experiments, that the later layers generate better task-specific human explainability. Our approach works with any score-generation method that generates layer-specific importance scores and requires no re-training. Most importantly, it shows that information in the later layers of the transformer stack are more important for model classification performance (The Hiding Game) as well as for human consistency (Token Overlap). In the future, we look to extend this work to tasks beyond classification. We also plan to further explore and leverage the geometric relationship between the feature embedding and token spaces first established in this paper. 

\subsubsection*{Acknowledgments}
This material is based upon work supported by the United States Air Force under Contract No. FA8650-19-C-6038 and FA8650-21-C-1168. Any opinions, findings and conclusions or recommendations expressed in this material are those of the author(s) and do not necessarily reflect the views of the Air Force or Department of Defense.

\bibliography{refs}
\bibliographystyle{icml2023}

\newpage
\appendix
\onecolumn
\section{Appendix} \label{app}

\subsection{Experimental Setup} \label{exp_setup}

\underline{Datasets:} 
\begin{enumerate}
\item SST-2 \cite{socher2013recursive}: This is a two class version of the Stanford sentiment analysis corpus where each sample is a full sentence from a movies review labeled as either Negative (class 0) or Positive (class 1) sentiment. It is split to have 67,349 training samples, 872 validation samples, and 1821 test samples; however the test labels are not publicly available and the validation set is commonly used for experiments in numerous paper including this one. 
\item AG News \cite{zhang2015character}: This is a four class version of a corpus collected by \cite{agnews} from over 2000 news sources. Each sample is a full sentence from a news article labeled as belonging to the World (class 0), Sports (class 1), Business (class 2), or Sci/Tech (class 3) topics. It is split to have 120,000 training samples and 7,600 test samples. 
\end{enumerate}

\underline{Models:} \\
AllenNLP Interpret on a RoBERTa large model \cite{Wallace2019AllenNLP}
\begin{enumerate} 
\item Simple \cite{simonyan2014deep}: gradient of the loss with respect to each token normalized by the $\ell_1$ norm
\item Smooth \cite{smilkov2017smoothgrad}: average the gradient over noisy input sequences (add white noise to embeddings)
\item Integrated \cite{sundararajan2017axiomatic}: integrating the gradient with 10 samples along the path from an embedding of all zeros to the original input sequence
\end{enumerate}
ThermoStat on a RoBERTa base model \cite{feldhus2021ThermoStat}
\begin{enumerate}
\item GRADxACT: simple element-wise product of gradient and activation
\item Integrated \cite{sundararajan2017axiomatic}: same as above, except 25 samples along the path
\item LIME \cite{ribeiro2016should}: sample 25 points around input sequence and use predictions at sample points to train a simpler interpretable model
\item Occlusion \cite{zeiler2014visualizing}: perturbation based approach, replace sliding window (3 tokens) with baseline and compute difference in prediction
\item Shapley \cite{castro2009polynomial}: add a random permutation of tokens from the input sequence to a baseline, look at difference in prediction after each addition, perform 25 times and average over them
\item DeepLiftShap \cite{lundberg2017unified}: approximates Shapely values, computes DeepLift attributions for each input-baseline pair, average over baselines
\item GradientShap \cite{lundberg2017unified}: approximates Shapely values, computes the expectations of gradients by randomly sampling 5 times from the distribution of baselines
\end{enumerate}
Decoded Grad-CAM layers implemented on RoBERTa base from HuggingFace \cite{wolf2020transformers} 

\underline{Metrics:}
\begin{enumerate}
\item The Hiding / Revealing game: Order tokens in descending amounts of importance for each method, hiding each token one by one.  Methods with a better grasp of importance will reduce the prediction accuracy of the network faster by hiding tokens that really matter.  The Revealing Game is the converse, which slowly reveals important tokens. This method was used in \cite{fong2017interpretable, castanon2018visualizing}.
\item Percentage of Token Overlaps: For every pair of classes we count the number of tokens in the top $k$ most-important for both classes and divide total count by $\binom{C}{2} * k $ where $C$ is number of classes.  This yields a measure of how unique the tokens we believe are important for identify a class are.
\end{enumerate}

\subsection{Additional Hiding / Revealing Game Results} \label{all_layers}

\begin{figure}[H]
\centering
\label{fig:sst_all}
\begin{tabular}{c c@{}c}
& \underline{Even Layers} & \underline{Odd Layers} \\
\rotatebox[origin=l]{90}{\hspace*{1.2cm} Revealing Game} &
\includegraphics[width=0.45\linewidth]{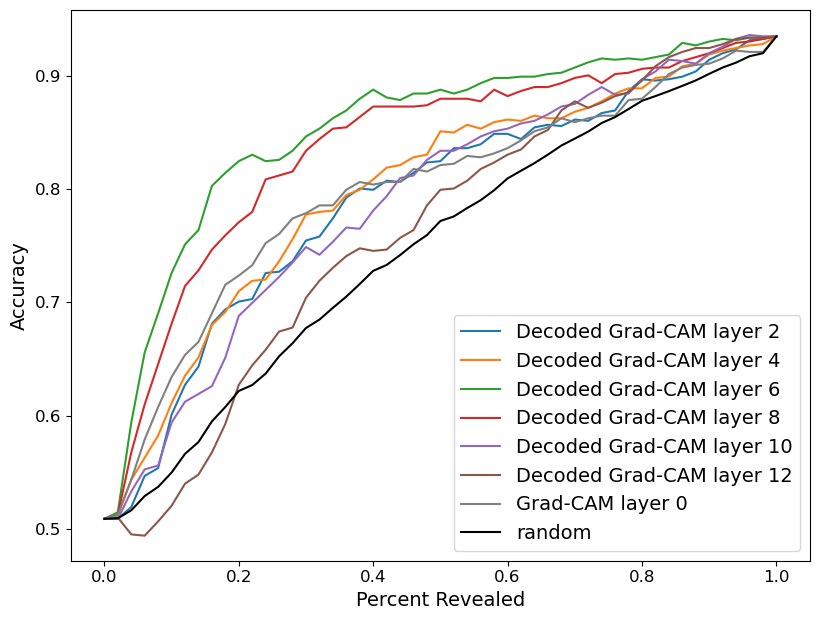} &
\includegraphics[width=0.45\linewidth]{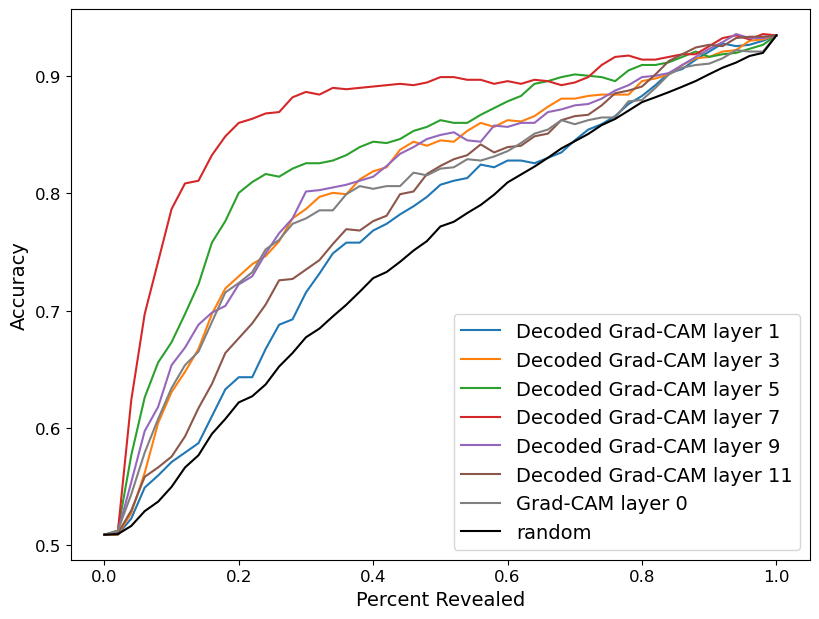} \\
\rotatebox[origin=l]{90}{\hspace*{1.2cm} Hiding Game} &
\includegraphics[width=0.45\linewidth]{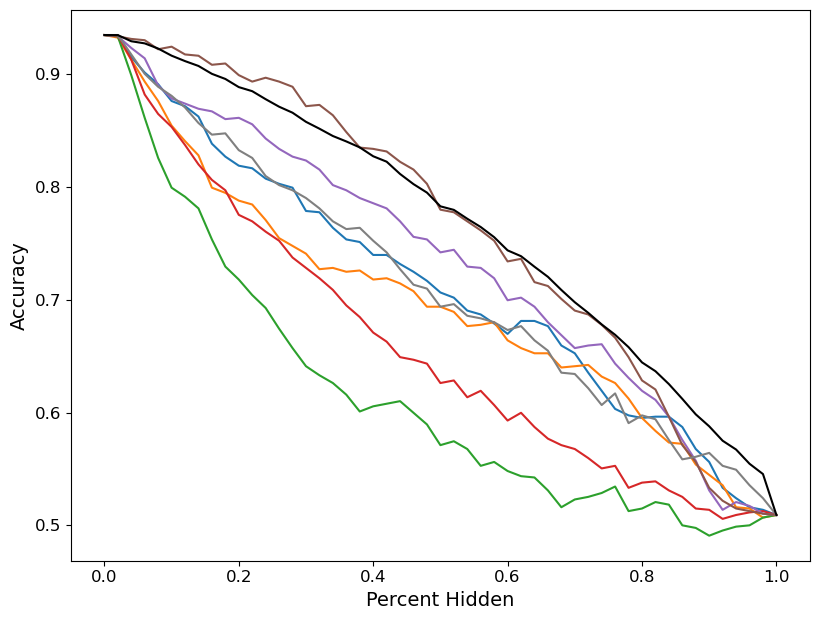} &
\includegraphics[width=0.45\linewidth]{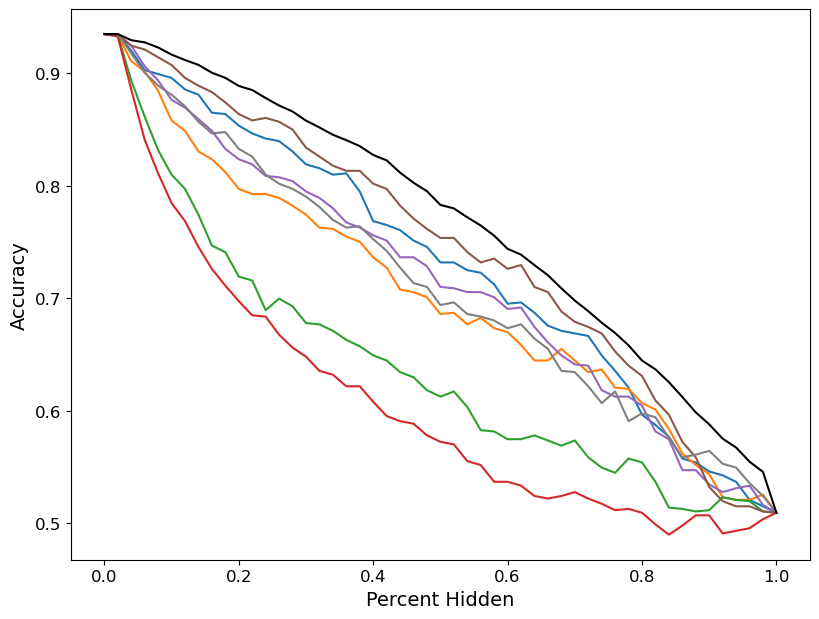} \\
\end{tabular}
\caption{All Decoded Grad-CAM layers on the SST-2 dataset.}
\end{figure}

\begin{figure}[H]
\centering
\label{fig:ag_all}
\begin{tabular}{c c@{}c}
& \underline{Even Layers} & \underline{Odd Layers} \\
\rotatebox[origin=l]{90}{\hspace*{1.2cm} Revealing Game} &
\includegraphics[width=0.45\linewidth]{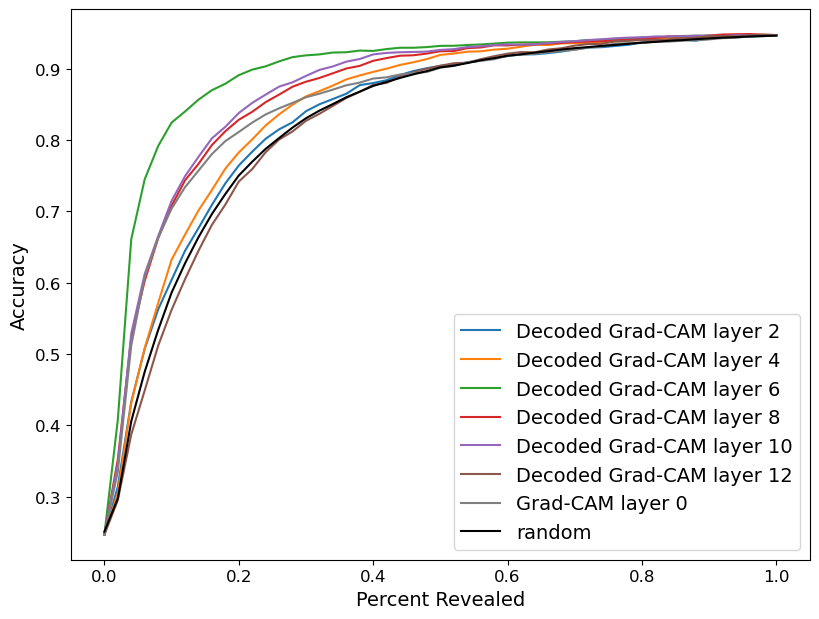} &
\includegraphics[width=0.45\linewidth]{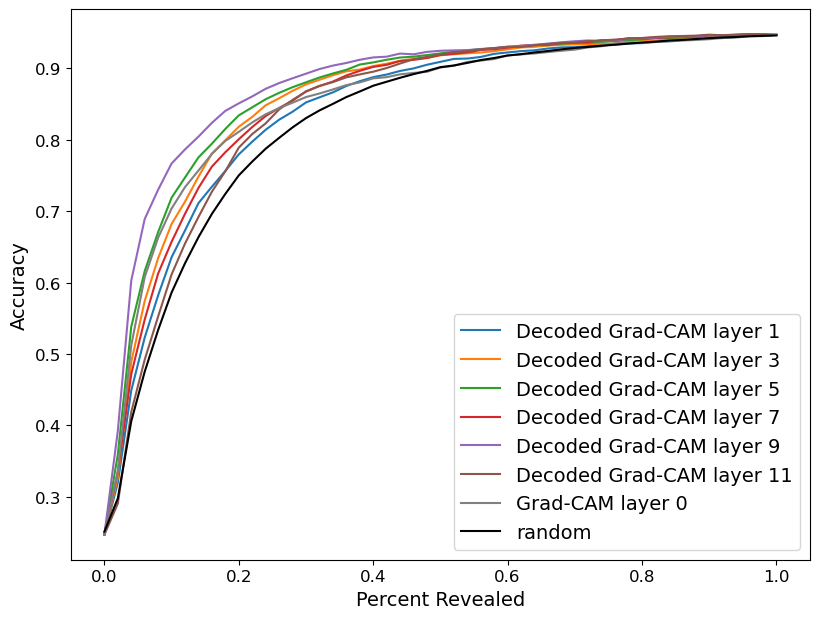} \\
\rotatebox[origin=l]{90}{\hspace*{1.2cm} Hiding Game} &
\includegraphics[width=0.45\linewidth]{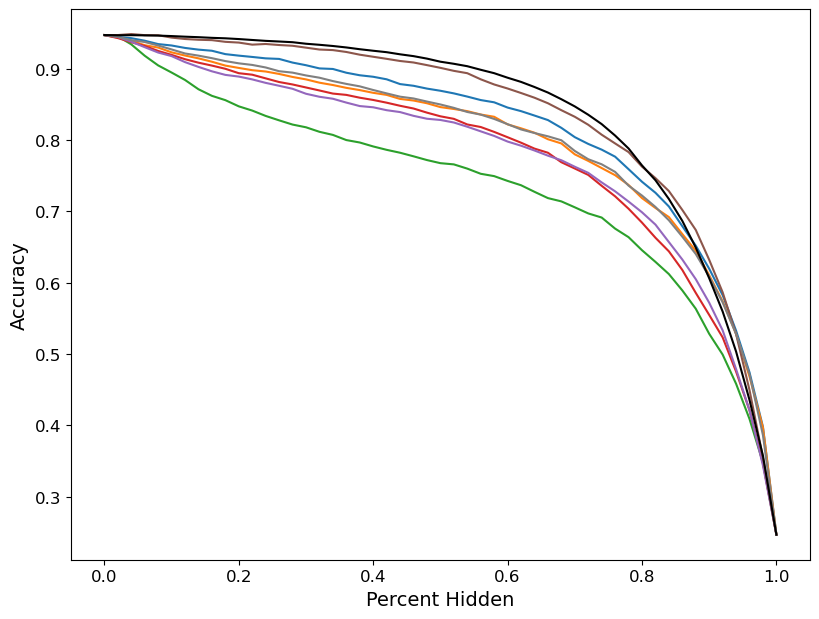} &
\includegraphics[width=0.45\linewidth]{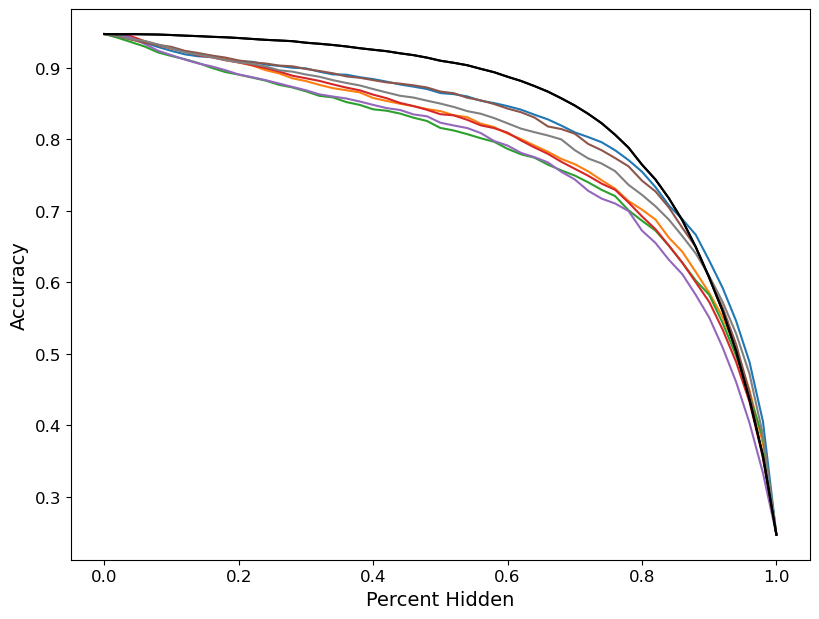} \\
\end{tabular}
\caption{All Decoded Grad-CAM layers on the AG News dataset.}
\end{figure}

\subsubsection{Other Saliency Score Results} \label{game_other_results}

In order to show generalizability of our method described in Algorithm \ref{alg} to more than Grad-CAM, we also show results replacing Grad-CAM with the “Simple” method from AllenNLP Interpret. Below we show the performance of layers 5, 6, and 7 using our decoding method versus the original “Simple” method. Note: there is a slight difference in performance due to the AllenNLP Interpret using a RoBERTa-large model and our method using a RoBERTa-base model.  

Similar to the Grad-CAM results, only including gradients above certain middle layers for the “Simple” method can have better performance than including all gradients by backpropagating to the embedding layer. 

\begin{figure}[H]
\centering
\begin{subfigure}{0.49\textwidth}
    \centering
    \includegraphics[width=\linewidth]{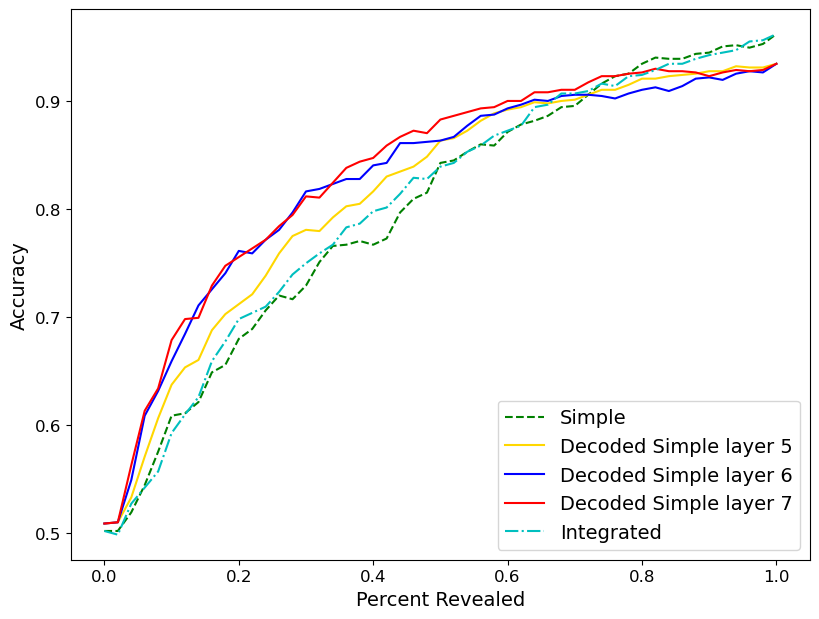}
    \caption{Revealing Game}
\end{subfigure}
\hfill
\begin{subfigure}{0.49\textwidth}
    \centering
    \includegraphics[width=\linewidth]{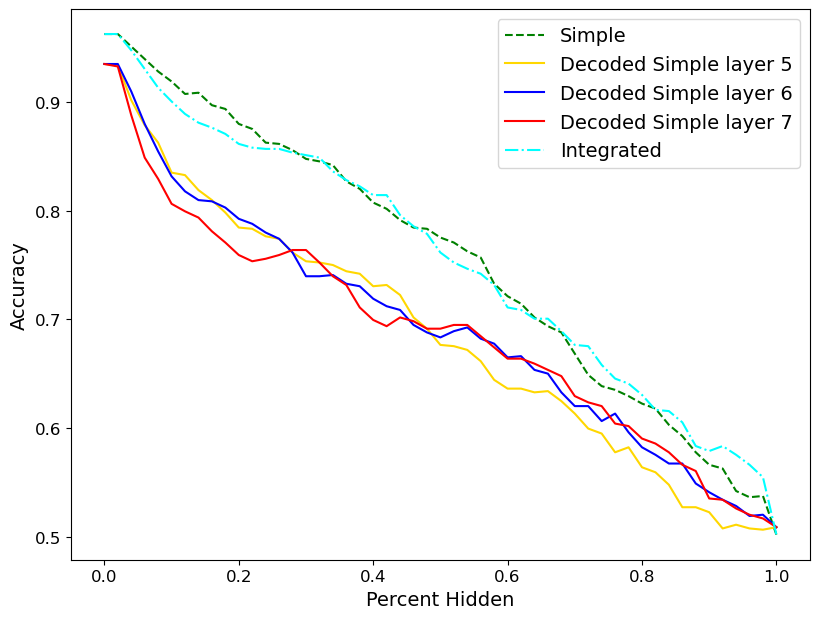}
    \caption{Hiding Game}
\end{subfigure} 
\caption{SST-2 binary sentiment classification dataset.}
\label{sst_hiding_game_simple}
\end{figure}

\begin{figure}[H]
\centering
\begin{subfigure}{0.49\textwidth}
    \centering
    \includegraphics[width=\linewidth]{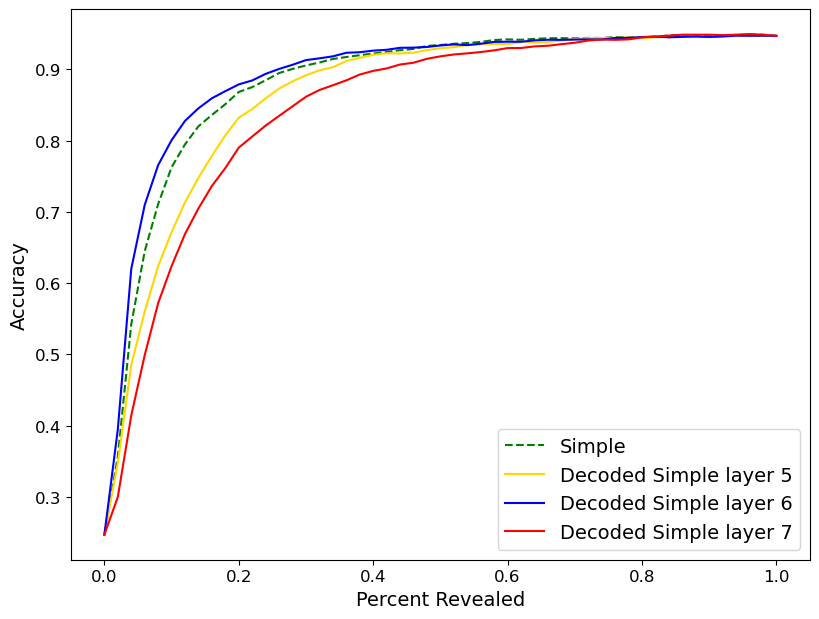}
    \caption{Revealing Game}
\end{subfigure}
\hfill
\begin{subfigure}{0.49\textwidth}
    \centering
    \includegraphics[width=\linewidth]{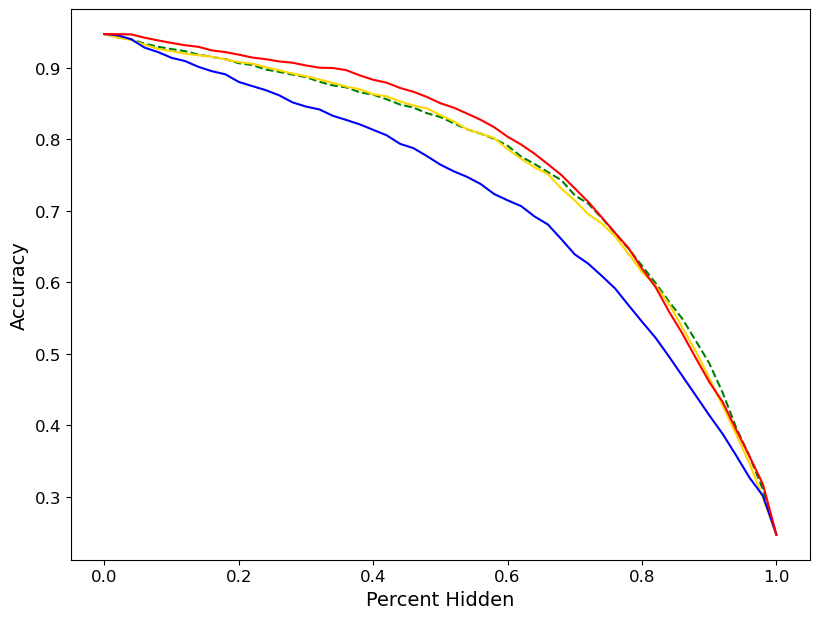}
    \caption{Hiding Game}
\end{subfigure} 
\caption{AG News four topic classification dataset.}
\label{agnews_hiding_game_simple}
\end{figure}

Additional results for Table~\ref{auc}.
\begin{table}[H]
\caption{\underline{Area Under the Curve for Revealing Game ($\uparrow$ is better) and Hiding Game ($\downarrow$ is better)}}
\vskip\baselineskip
\begin{minipage}{0.5\linewidth}
\centering
\begin{tabular}{l|c|c}
Explainer & Revealing & Hiding \\ \hline 
& & \\
Simple & 0.799 & 0.756 \\
Decoded Simple $l_5$ & 0.816 & \textbf{0.687} \\
Decoded Simple $l_6$ & 0.828 & 0.694 \\
Decoded Simple $l_7$ & \textbf{0.835} & 0.689 \\
\end{tabular}
\subcaption{SST-2 Dataset (Figure~\ref{sst_hiding_game_simple})}
\end{minipage}
\hfill
\begin{minipage}{0.5\linewidth}
\centering
\begin{tabular}{l|c|c}
Explainer & Revealing & Hiding \\ \hline 
& & \\
Simple $l_0$ & 0.883 & 0.765 \\
Decoded Simple $l_5$ & 0.866 & 0.762 \\
Decoded Simple $l_6$ & \textbf{0.891} & \textbf{0.716} \\
Decoded Simple $l_7$ & 0.847 & 0.774 \\
\end{tabular}
\subcaption{AG News Dataset (Figure~\ref{agnews_hiding_game_simple})}
\end{minipage}
\end{table}

\subsection{Word Clouds} \label{word_clouds}

The size of the tokens in the word clouds are reflective of the weighted tokens scores.

\begin{figure}[H]
\begin{tabular}{c c@{}c}
& \underline{Predicted Negative} & \underline{Predicted Positive} \\
\\
\rotatebox[origin=l]{90}{\hspace*{1cm} Simple} &
\includegraphics[width=0.45\linewidth]{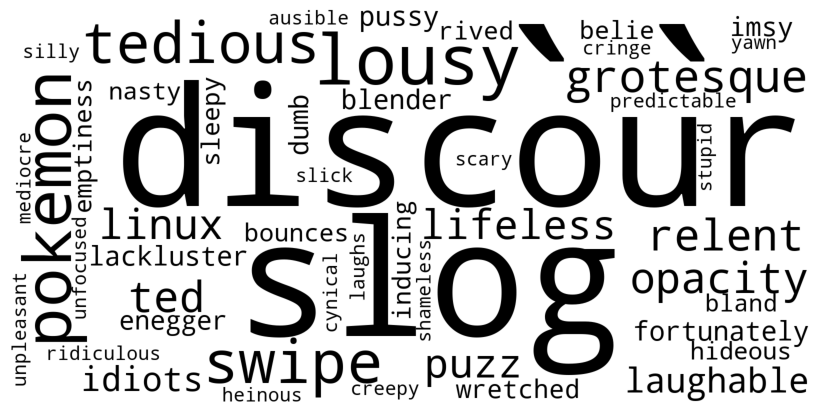} &
\includegraphics[width=0.45\linewidth]{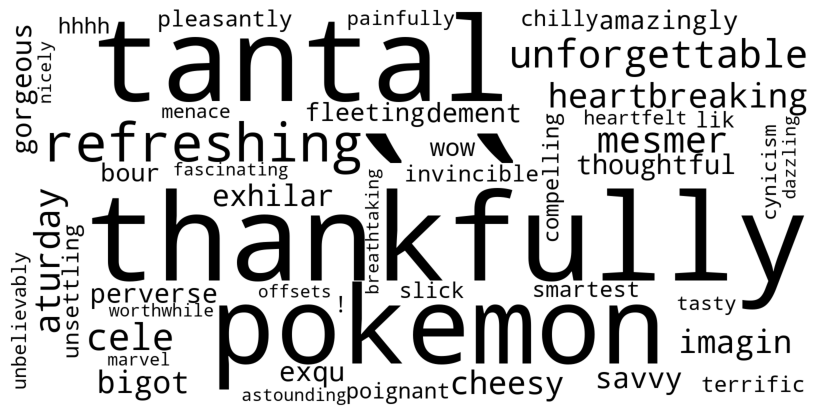} \\
\rotatebox[origin=l]{90}{\hspace*{1cm} Smooth} &
\includegraphics[width=0.45\linewidth]{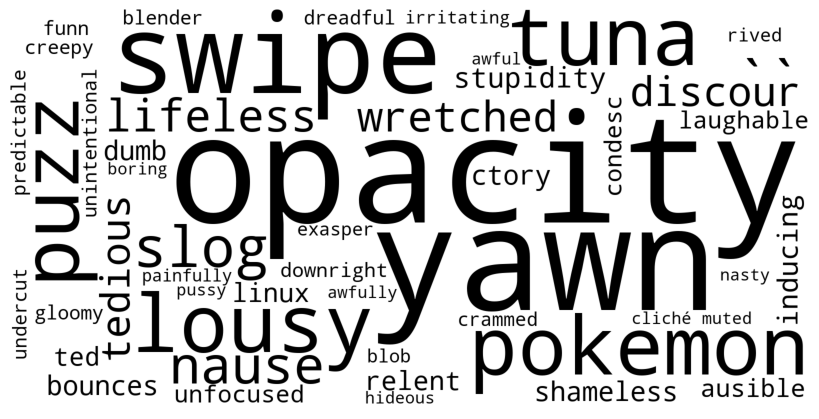} & 
\includegraphics[width=0.45\linewidth]{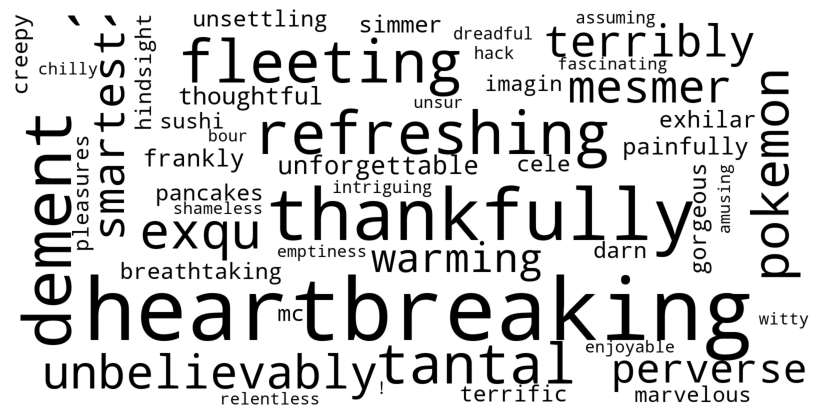} \\
\rotatebox[origin=l]{90}{\hspace*{1cm} Integrated} &
\includegraphics[width=0.45\linewidth]{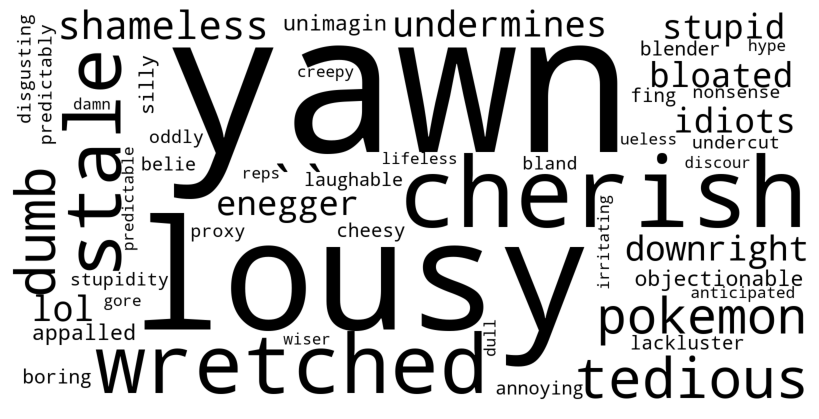} &
\includegraphics[width=0.45\linewidth]{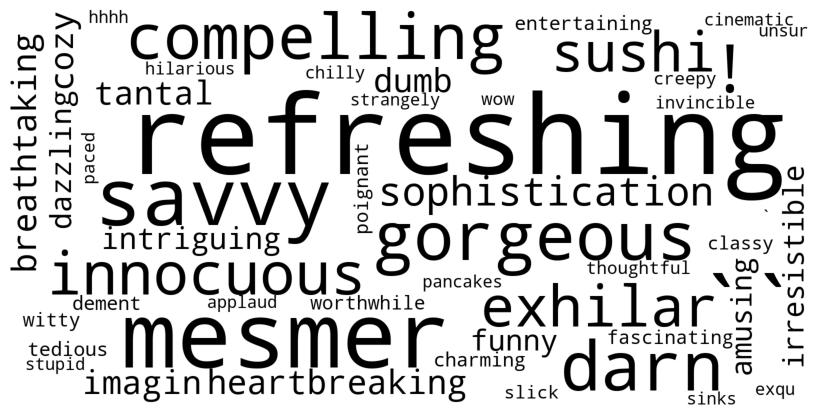} \\
\rotatebox[origin=l]{90}{\hspace*{0.05cm} Decoded Grad-CAM $l_7$} &
\includegraphics[width=0.45\linewidth]{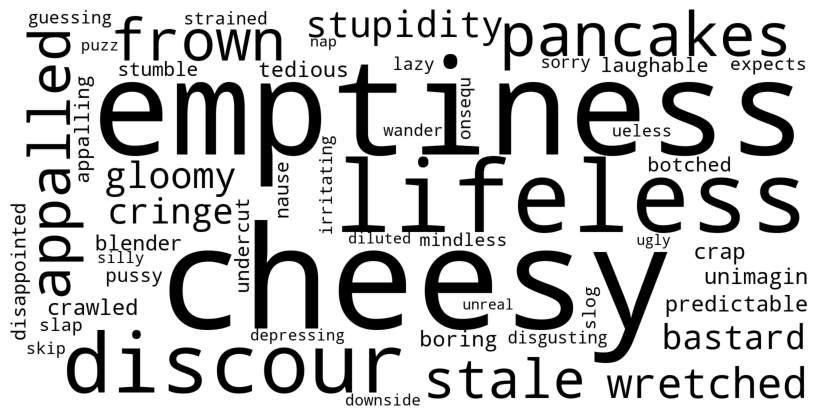} &
\includegraphics[width=0.45\linewidth]{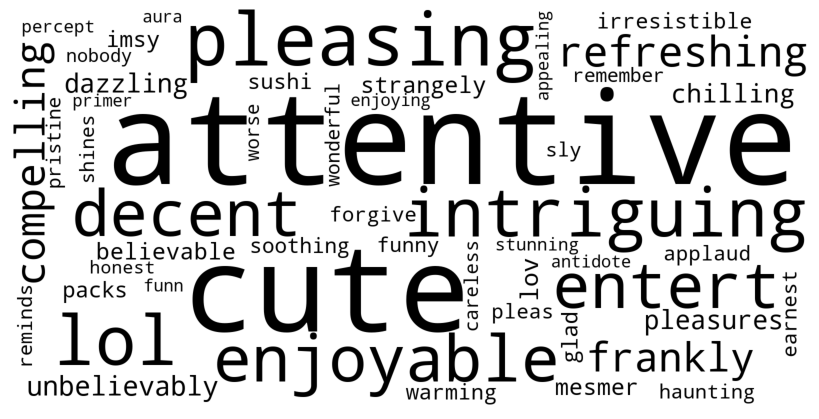}
\end{tabular}
\caption{Word Clouds with Top 50 Tokens for SST-2 dataset.}
\label{fig:sst2_word_clouds}
\end{figure}

 While the left column (Predicted Negative) of Figure~\ref{fig:sst2_word_clouds} is composed of generally negative terms, there are some more puzzling tokens that are deemed important according to the AllenNLP Interpret explainers such as ` ` and pokemon. The right column (Predicted Positive) also seems to have some tokens with negative connotations such as \{menace, dement\} in the Simple, \{terribly, dement, hack. dreadful\} in the Smooth, and \{dumb, stupid, tedious\} in the Integrated explainer. 

\begin{figure}[H]
\begin{tabular}{c c@{}c}
& \underline{Predicted Negative} & \underline{Predicted Positive} \\
\\
\rotatebox[origin=l]{90}{\hspace*{1cm} Decoded Simple $l_5$} &
\includegraphics[width=0.45\linewidth]{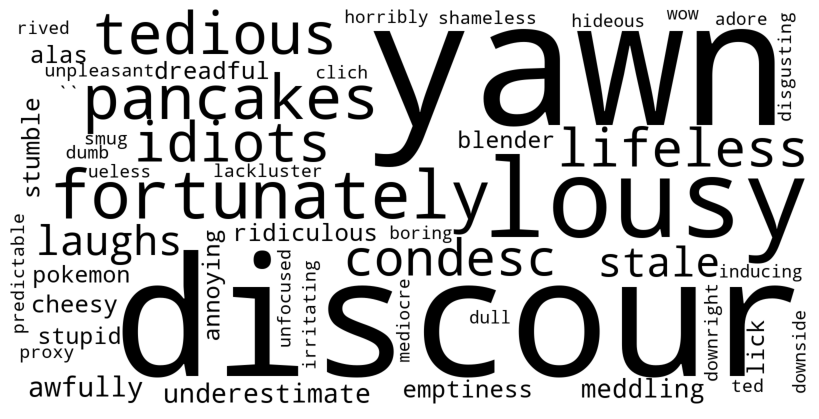} &
\includegraphics[width=0.45\linewidth]{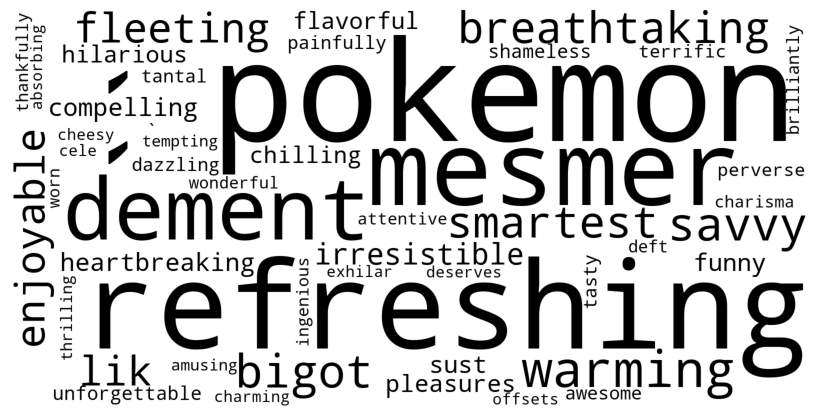} \\
\rotatebox[origin=l]{90}{\hspace*{1cm} Decoded Simple $l_6$} &
\includegraphics[width=0.45\linewidth]{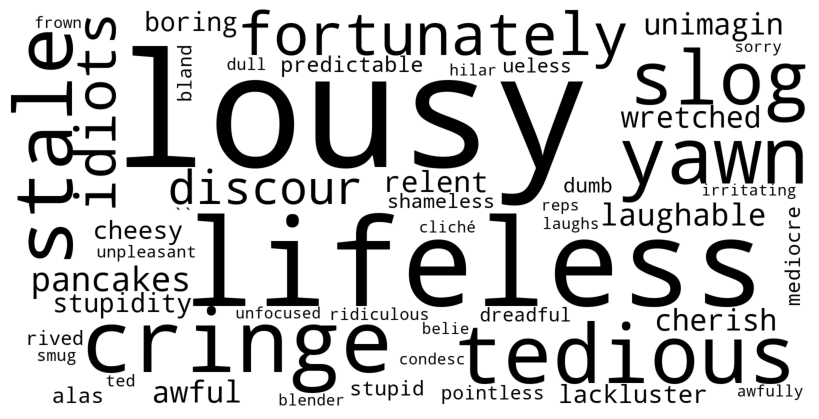} & 
\includegraphics[width=0.45\linewidth]{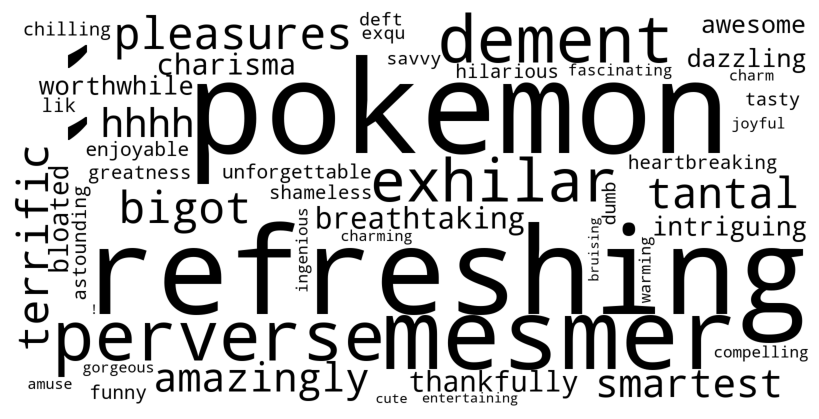} \\
\rotatebox[origin=l]{90}{\hspace*{1cm} Decoded Simple $l_7$} &
\includegraphics[width=0.45\linewidth]{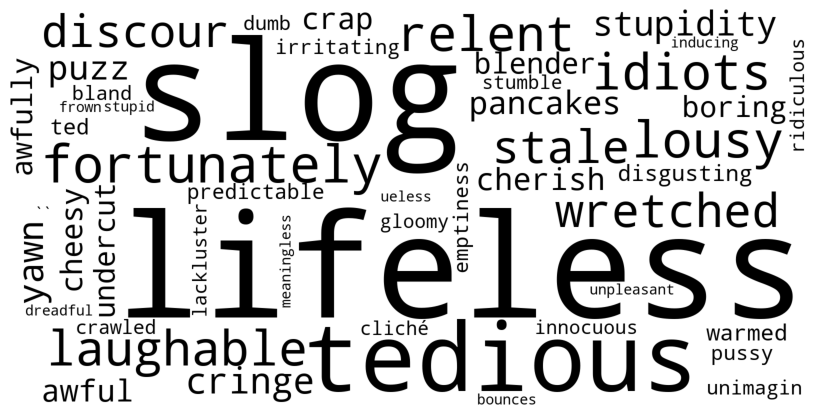} &
\includegraphics[width=0.45\linewidth]{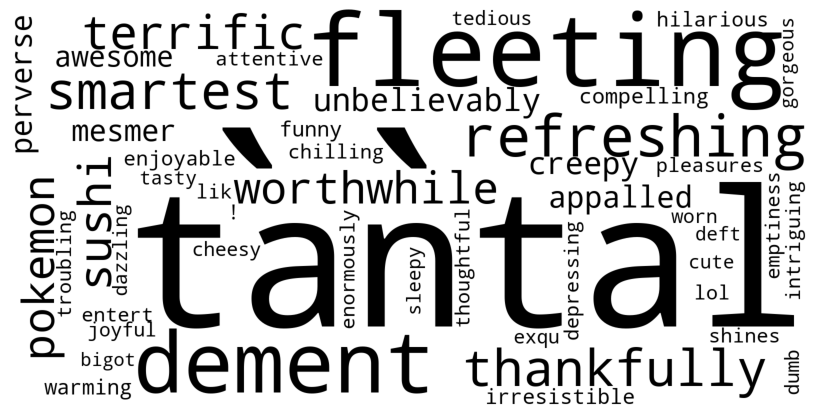} 
\end{tabular}
\caption{Word Clouds with Top 50 Tokens for SST-2 dataset.}
\label{fig:sst2_word_clouds_simple}
\end{figure}

\newpage

\begin{figure}[H]
\centering
\begin{tabular}{c@{}c}
\rotatebox[origin=c]{270}{Figure~\ref{fig:ag_news_word_clouds}: Word Clouds with Top 50 Tokens for AG News dataset} &
\begin{tabular}{c@{}c@{}c@{}c c}
Integrated & Shapely & Occlusion & Decoded Grad-CAM $l_6$ & \hfill \\
\includegraphics[angle=270,origin=c, width=0.17\linewidth]{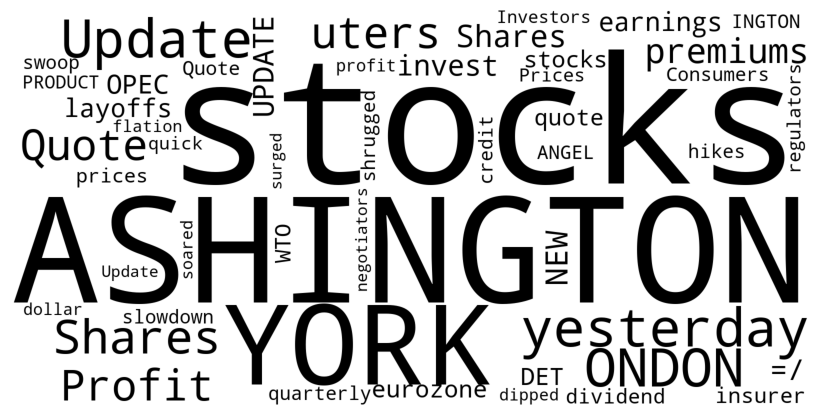} &
\includegraphics[angle=270,origin=c, width=0.17\linewidth]{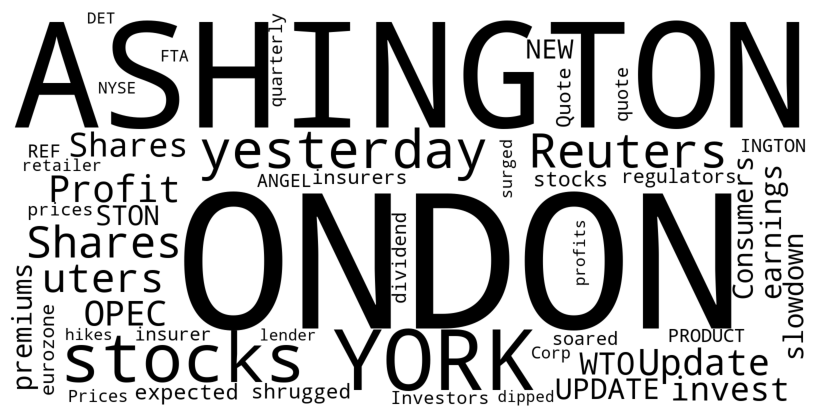} &
\includegraphics[angle=270,origin=c, width=0.17\linewidth]{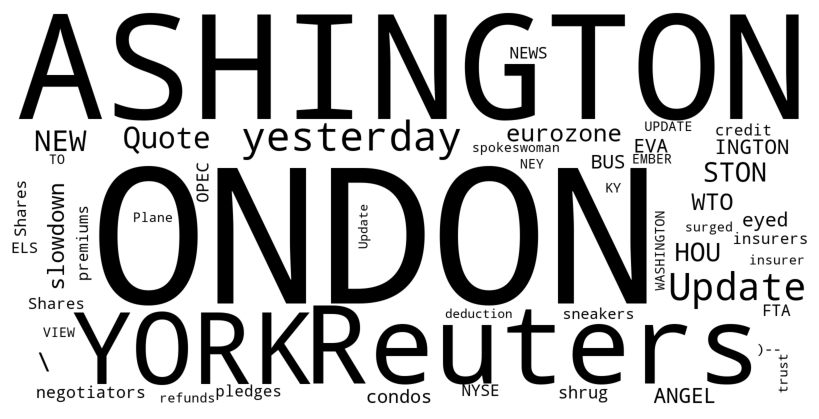} &
\includegraphics[angle=270,origin=c, width=0.17\linewidth]{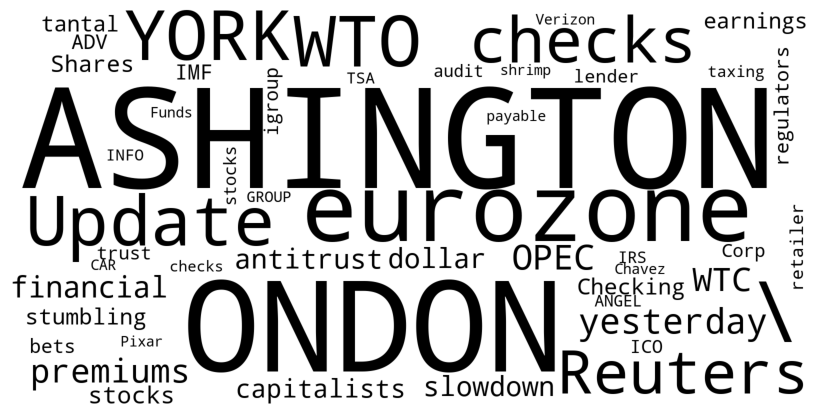} & 
\rotatebox[origin=c]{270}{\hspace*{-3cm} \underline{Business}} \\
\includegraphics[angle=270,origin=c, width=0.17\linewidth]{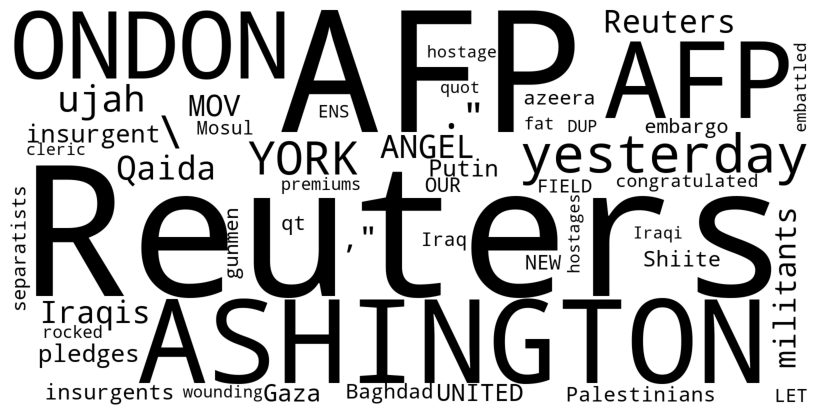} &
\includegraphics[angle=270,origin=c, width=0.17\linewidth]{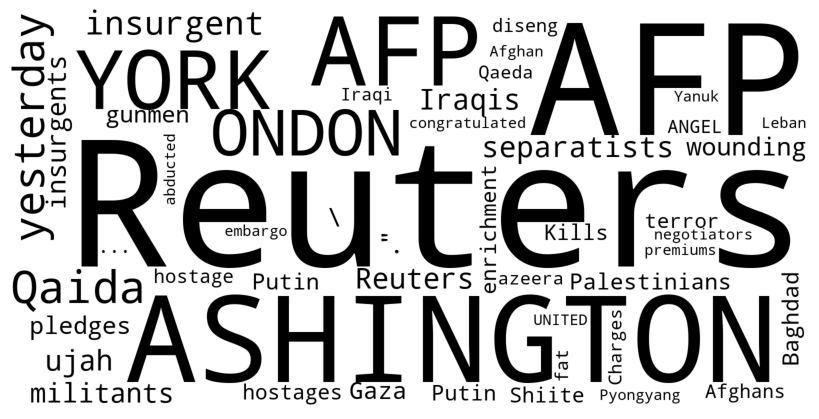} &
\includegraphics[angle=270,origin=c, width=0.17\linewidth]{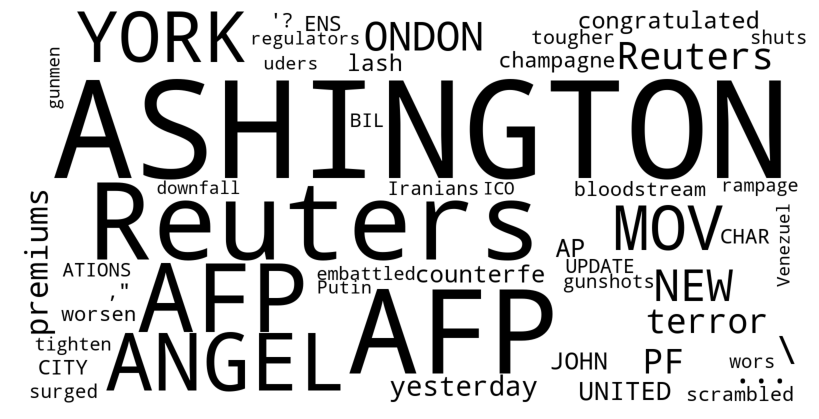} &
\includegraphics[angle=270,origin=c, width=0.17\linewidth]{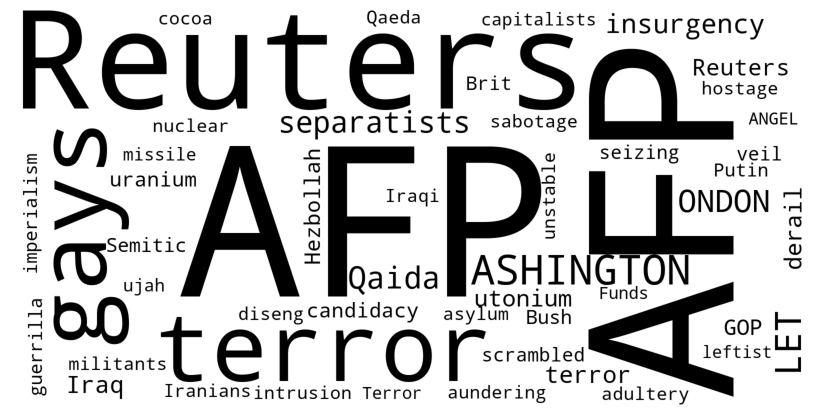} & 
\rotatebox[origin=c]{270}{\hspace*{-3cm} \underline{World}} \\
\includegraphics[angle=270,origin=c, width=0.17\linewidth]{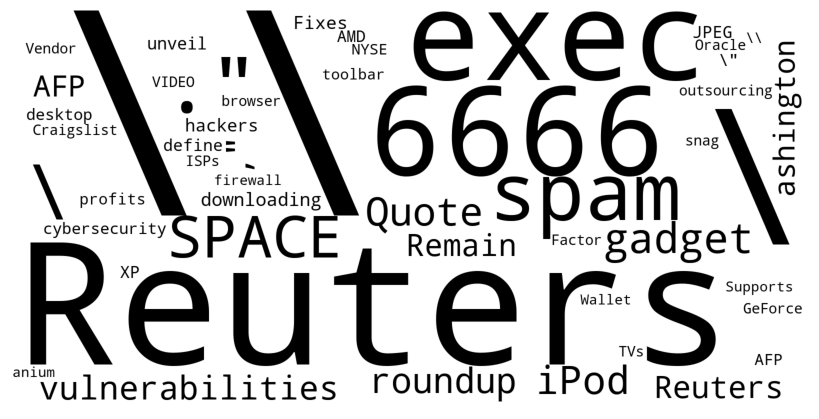} &
\includegraphics[angle=270,origin=c, width=0.17\linewidth]{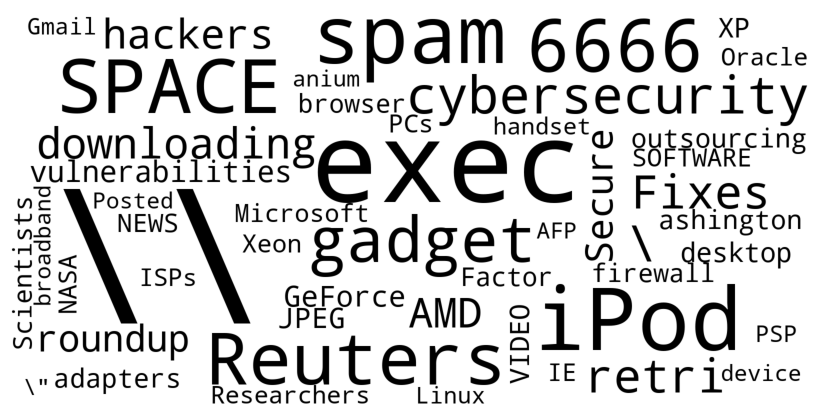} & 
\includegraphics[angle=270,origin=c, width=0.17\linewidth]{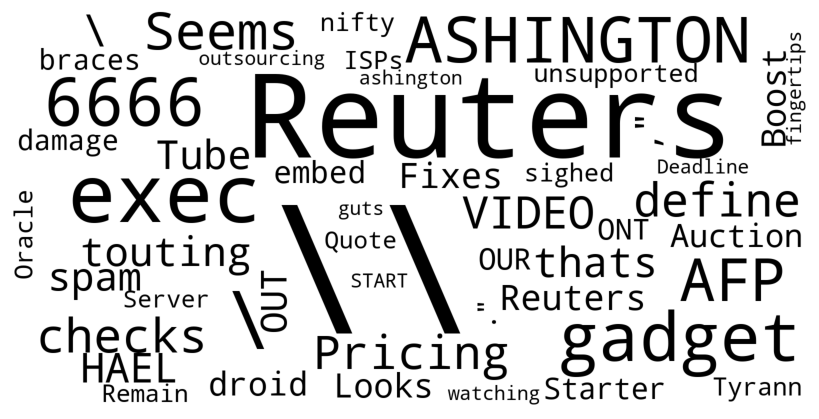} &
\includegraphics[angle=270,origin=c, width=0.17\linewidth]{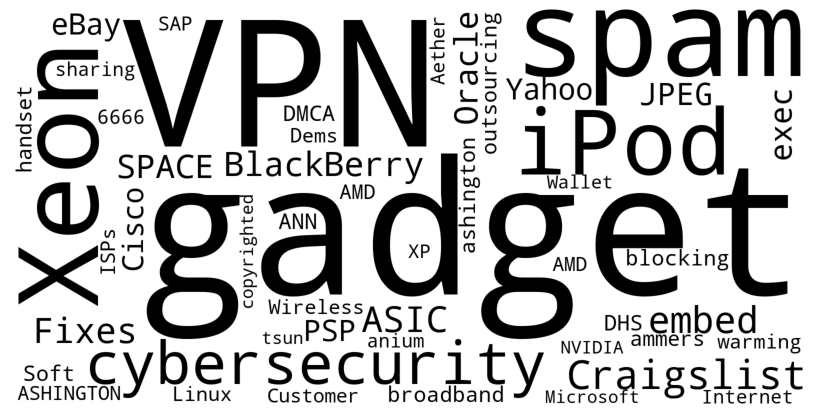} & 
\rotatebox[origin=c]{270}{\hspace*{-3cm} \underline{Science / Tech}} \\
\includegraphics[angle=270,origin=c, width=0.17\linewidth]{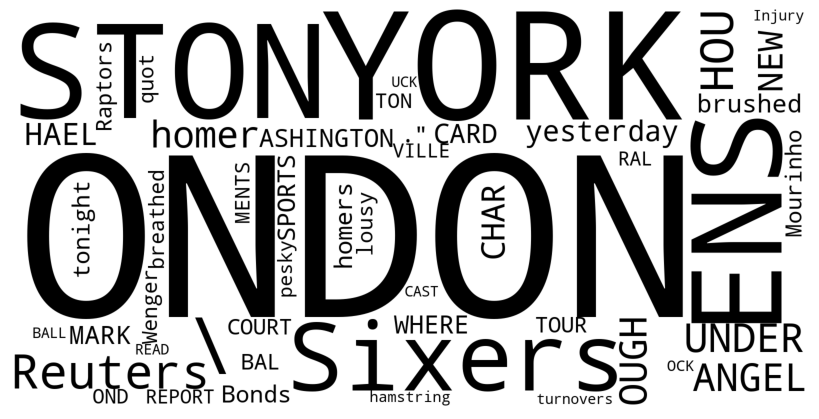} &
\includegraphics[angle=270,origin=c, width=0.17\linewidth]{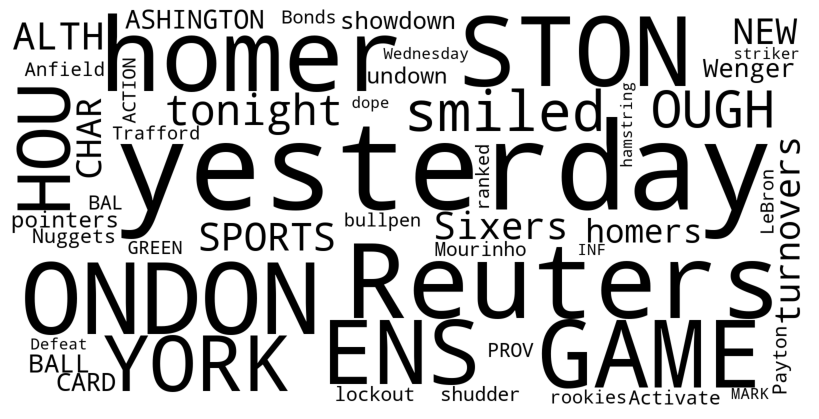} &
\includegraphics[angle=270,origin=c, width=0.17\linewidth]{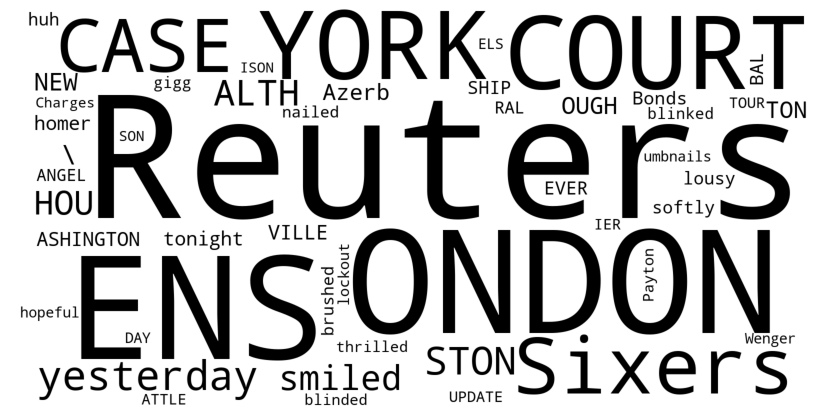} & 
\includegraphics[angle=270,origin=c, width=0.17\linewidth]{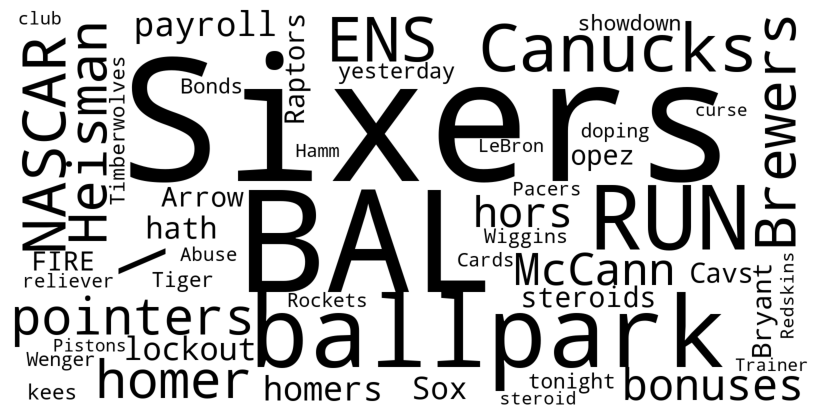} &
\rotatebox[origin=c]{270}{\hspace*{-3cm} \underline{Sports}} \\
\end{tabular}
\end{tabular}
\phantomcaption
\label{fig:ag_news_word_clouds}
\end{figure}

\newpage

\subsection{Overlapping Tokens} \label{overlap}

\begin{table}[H]
\caption{Tokens in Top 50 appearing in both World and Sports classes of AG News dataset}
\begin{center}
\begin{tabular}{l|l}
\multicolumn{1}{c}{\bf Explainer} \vline &\multicolumn{1}{c}{\bf Overlapping Tokens}
\\ \hline \\
Shapely & \begin{minipage}[t]{0.75\columnwidth}%
NEW,  apologised, FIELD,  Defeat,  CHAR, ONDON,  YORK,  UNITED,  Thursday,  shook, ASHINGTON,  Wednesday,  Tuesday, ENS,  tonight,  roared, Reuters,  AP,  ANGEL,  yesterday 
\end{minipage}\tabularnewline
\\
Occlusion & \begin{minipage}[t]{0.75\columnwidth}%
NEW, ELS,  Calif, AFP,  \textbackslash\textbackslash,  YORK, ONDON, ASHINGTON,  Tuesday, ENS, IJ,  UPDATE,  Charges,  awaits, IGH, ,,  Monday,  roared, Reuters, VER,  AP,  ANGEL,  yesterday,  BE 
\end{minipage}\tabularnewline
\\
Integrated & \begin{minipage}[t]{0.75\columnwidth}%
NEW,  apologised, FIELD,  Calif,  chilly,  \textbackslash\textbackslash, ONDON,  YORK,  Talks,  UNITED,  Thursday, ASHINGTON,  Wednesday, ENS,  PARK, Update,  embattled,  quot,  roared, Reuters,  AP, ANG, .,  ANGEL,  yesterday 
\end{minipage}\tabularnewline
\\
Decoded Grad-CAM $l_6$ & (none) \\
\end{tabular}
\end{center}
\end{table}

\begin{table}[H]
\caption{Tokens in Top 50 appearing in both World and Business classes of AG News dataset}
\begin{center}
\begin{tabular}{l|l}
\multicolumn{1}{c}{\bf Explainer} \vline &\multicolumn{1}{c}{\bf Overlapping Tokens}
\\ \hline \\
Shapely & \begin{minipage}[t]{0.75\columnwidth}%
 NEW, ONDON,  YORK,  Charges, expected,  Thursday,  rattled, ASHINGTON, Reuters,  Wednesday,  premiums,  Tuesday,  plunged,  surged,  ANGEL,  yesterday \end{minipage}\tabularnewline
\\
Occlusion & \begin{minipage}[t]{0.75\columnwidth}%
EVA,  NEW, ELS,  Calif,  optimism, \textbackslash\textbackslash, ONDON,  YORK, ASHINGTON,  Tuesday,  UPDATE,  Charges, Update,  Shares,  hammered,  Monday,  surged,  embattled, Reuters,  MOV,  premiums,  Profit,  regulators,  ANGEL,  yesterday \end{minipage}\tabularnewline
\\
Integrated & \begin{minipage}[t]{0.75\columnwidth}%
 NEW,  negotiators,  YORK, ONDON,  pledges,  Thursday, ASHINGTON,  Wednesday,  Tuesday,  plunged,  tighten, Update,  IPO,  soared,  surged,  Says, expected,  yesterday,  premiums,  ANGEL,  Charges \end{minipage}\tabularnewline
\\
Decoded Grad-CAM $l_6$ & \begin{minipage}[t]{0.75\columnwidth}%
ONDON,  Funds, ASHINGTON, Reuters,  capitalists,  ANGEL \end{minipage}\tabularnewline
\end{tabular}
\end{center}
\end{table}

\begin{table}[H]
\caption{Tokens in Top 50 appearing in both Sports and Business classes of AG News dataset}
\begin{center}
\begin{tabular}{l|l}
\multicolumn{1}{c}{\bf Explainer} \vline &\multicolumn{1}{c}{\bf Overlapping Tokens}
\\ \hline \\
Shapely & \begin{minipage}[t]{0.75\columnwidth}%
 NEW, NEY, STON, ONDON,  YORK,  Thursday, ASHINGTON,  Wednesday,  Tuesday, INGTON,  eased, OND,  woes,  HOU, Reuters,  ANGEL,  MARK,  yesterday,  rallied \end{minipage}\tabularnewline
\\
Occlusion & \begin{minipage}[t]{0.75\columnwidth}%
 NEW, OCK, BUR, ELS, STON,  Calif,  \textbackslash\textbackslash, ONDON,  YORK,  roaring,  Thursday, ASHINGTON,  Wednesday, TON,  Tuesday, INGTON, STER,  eased,  UPDATE, ANC,  Charges,  --,  Monday,  HOU,  bruised, Reuters,  NEWS, BUS,  ANGEL,  yesterday \end{minipage}\tabularnewline
\\
Integrated & \begin{minipage}[t]{0.75\columnwidth}%
ANC,  NEW, OCK,  YORK, ONDON,  HOU,  Thursday, Update, ASHINGTON,  Wednesday, INGTON, STER,  TOR,  ANGEL,  MARK,  yesterday,  rallied \end{minipage}\tabularnewline
\\
Decoded Grad-CAM $l_6$ & \textbackslash\textbackslash,  yesterday
\end{tabular}
\end{center}
\end{table}

\begin{table}[H]
\caption{Tokens in Top 50 appearing in both World and Sci/Tech classes of AG News dataset}
\begin{center}
\begin{tabular}{l|l}
\multicolumn{1}{c}{\bf Explainer} \vline &\multicolumn{1}{c}{\bf Overlapping Tokens}
\\ \hline \\
Shapely & \begin{minipage}[t]{0.75\columnwidth}%
NEW, ONDON,  YORK,  Charges, expected,  Thursday,  rattled, ASHINGTON, Reuters,  Wednesday,  premiums,  Tuesday,  plunged,  surged,  ANGEL,  yesterday
\end{minipage}\tabularnewline
\\
Occlusion & \begin{minipage}[t]{0.75\columnwidth}%
expected, Update, ASHINGTON, ,",  Reuters, terror, Reuters, AFP,  Calls, \textbackslash\textbackslash \end{minipage}\tabularnewline
\\
Integrated & \begin{minipage}[t]{0.75\columnwidth}%
 unveil,  AFP, Reuters, ,",  Reuters, AFP, .",  ANGEL, \textbackslash\textbackslash \end{minipage}\tabularnewline
\\
Decoded Grad-CAM $l_6$ & ASHINGTON
\end{tabular}
\end{center}
\end{table}

\begin{table}[H]
\caption{Tokens in Top 50 appearing in both Sports and Sci/Tech classes of AG News dataset}
\begin{center}
\begin{tabular}{l|l}
\multicolumn{1}{c}{\bf Explainer} \vline &\multicolumn{1}{c}{\bf Overlapping Tokens}
\\ \hline \\
Shapely & \begin{minipage}[t]{0.75\columnwidth}%
 showdown, Reuters,  HAS \end{minipage}\tabularnewline
\\
Occlusion & \begin{minipage}[t]{0.75\columnwidth}%
 ATT,  Boost,  --, ASHINGTON, ,", Reuters, HAEL,  WITH, STON, AFP, STER,  sighed, Factor, \textbackslash\textbackslash \end{minipage}\tabularnewline
\\
Integrated & \begin{minipage}[t]{0.75\columnwidth}%
 showdown, Reuters, HAEL,  Adds, .",  pesky,  ANGEL, \textbackslash\textbackslash \end{minipage}\tabularnewline
\\
Decoded Grad-CAM $l_6$ & (none)
\end{tabular}
\end{center}
\end{table}

\begin{table}[H]
\caption{Tokens in Top 50 appearing in both Business and Sci/Tech classes of AG News dataset}
\begin{center}
\begin{tabular}{l|l}
\multicolumn{1}{c}{\bf Explainer} \vline &\multicolumn{1}{c}{\bf Overlapping Tokens}
\\ \hline \\
Shapely & \begin{minipage}[t]{0.75\columnwidth}%
Quote,  Boost, Reuters, trust \end{minipage}\tabularnewline
\\
Occlusion & \begin{minipage}[t]{0.75\columnwidth}%
NEY, Update,  --,  quot,  Quote,  Consumers, ASHINGTON,  HERE, Reuters, STON, STER,  BlackBerry,  Customers, EMBER,  Pact,  \textbackslash\textbackslash,  Update \end{minipage}\tabularnewline
\\
Integrated & \begin{minipage}[t]{0.75\columnwidth}%
daq, Quote,  Craigslist, checks, ?, uters,  ANGEL,  profits,  Update \end{minipage}\tabularnewline
\\
Decoded Grad-CAM $l_6$ & ASHINGTON
\end{tabular}
\end{center}
\end{table}

\begin{table}[H]
\caption{Tokens in Top 50 appearing in both classes of SST-2 dataset}
\begin{center}
\begin{tabular}{l|l}
\multicolumn{1}{c}{\bf Explainer} \vline &\multicolumn{1}{c}{\bf Overlapping Tokens}
\\ \hline \\
Simple & slick, ` `, pokemon \\
Smooth & pokemon, creepy, shameless, dreadful, painfully, ` `  \\
Integrated & dumb, creepy, stupid, tedious, ` ` \\
Decoded Grad-CAM $l_7$ & (none) \\
\end{tabular}
\end{center}
\end{table}

\begin{figure}[H]
\centering
\begin{subfigure}{0.49\textwidth}
    \centering
    \includegraphics[width=\linewidth]{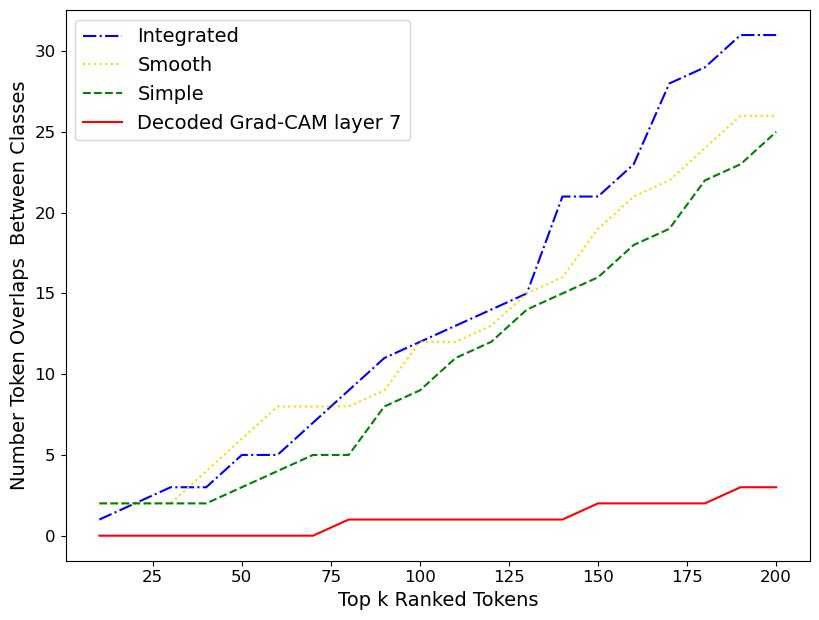}
\caption{SST-2 dataset}
\end{subfigure}
\hfill
\begin{subfigure}{0.49\textwidth}
    \centering
    \includegraphics[width=\linewidth]{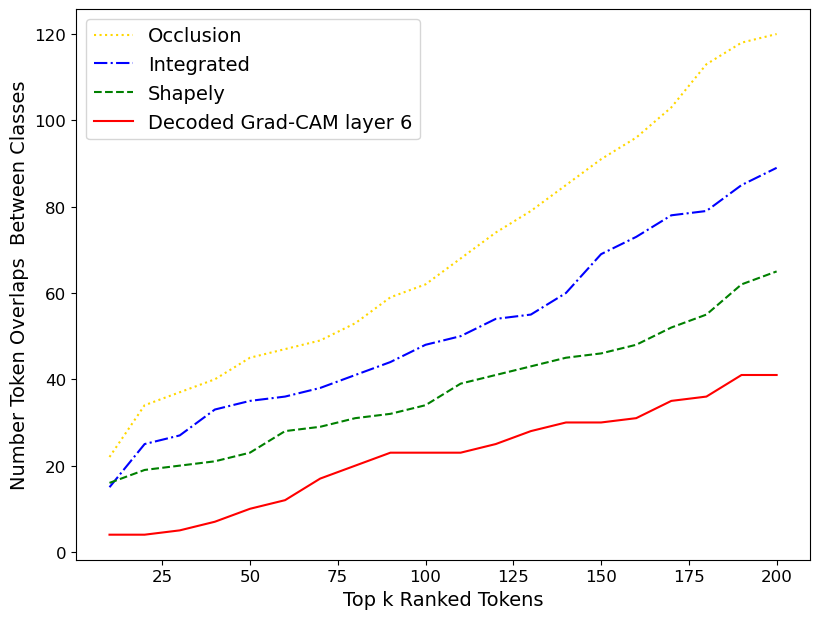}
\caption{AG News dataset}
\end{subfigure} 
\caption{Number of tokens that appear in multiple classes for the top $k$ most important tokens.}
\label{fig:tok_count_k}
\end{figure}

\newpage

\subsection{Examples of Highlighted Explanations} \label{highlights}

The underlined text above each snippet is the predicted class for that method with corresponding prediction probability and the intensity of the highlighted color reflects the relative importance of each token normalized for each input sequence. All snippets shown are of correctly predicted examples. We have provided highlights of all input sentences in the validation / test sets for both datasets in an attached supplementary file.

\vspace*{40pt}

\begin{figure}[H]
\begin{tabular}{m{2cm} | m{4cm} | m{2cm}}
Simple & \includegraphics[height=22pt]{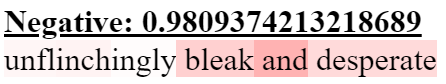} & \includegraphics[height=30pt]{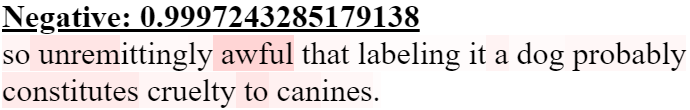} \\
Smooth & \includegraphics[height=21pt]{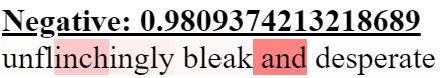} & \includegraphics[height=30pt]{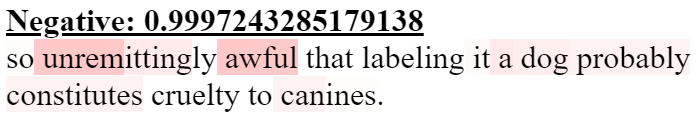} \\
Integrated & \includegraphics[height=22pt]{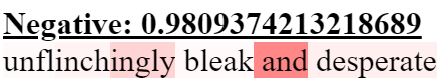} & \includegraphics[height=30pt]{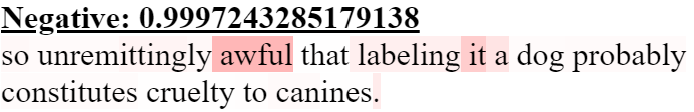} \\
Decoded Grad-CAM $l_7$ & \includegraphics[height=22pt]{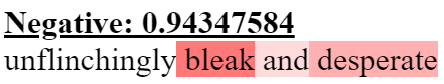} & \includegraphics[height=30pt]{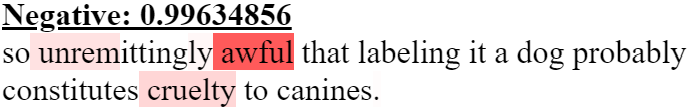}
\end{tabular}
\caption*{STT-2 Example 1: For the left column example, Decoded Grad-CAM $l_7$ and to an extent AllenNLP Interpret's Simple highlight the negative sentiment words "bleak" and "desperate", but all three of AllenNLP Interpret's methods also focus on "and". For the right column example, all four methods focus on "awful" and "unrem"(ittingly), but the AllenNLP Interpret's methods are more noisy with highlights on unrelated terms such as "it", "dog" and "constitutes".}
\end{figure}

\vspace*{60pt}

\begin{figure}[H]
\begin{tabular}{m{2cm} | m{5.6cm} | m{2cm}}
Simple & \includegraphics[height=30pt]{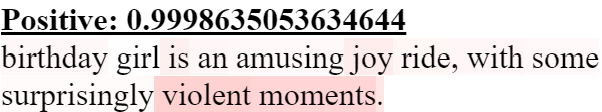} & \includegraphics[height=30pt]{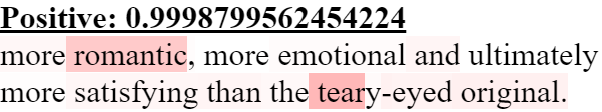} \\
Smooth & \includegraphics[height=30pt]{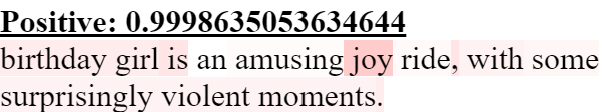} & \includegraphics[height=30pt]{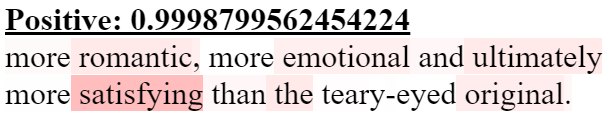} \\
Integrated & \includegraphics[height=30pt]{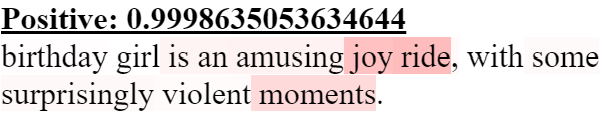} & \includegraphics[height=30pt]{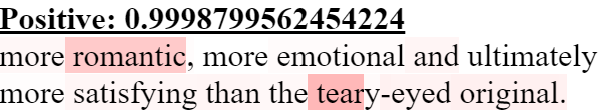}  \\
Decoded Grad-CAM $l_7$ & \includegraphics[height=30pt]{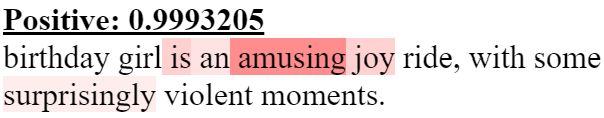} & \includegraphics[height=30pt]{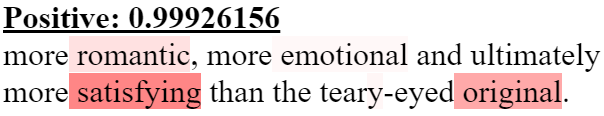} 
\end{tabular}
\caption*{STT-2 Example 2:  For the left column example, Decoded Grad-CAM $l_7$ focuses strongly on the positive phrases "is an amusing joy" and "surprising". The other methods also highlight these terms, but less clearly with unrelated words such as "moment" and potentially negative words such as "violent". For the right column example, all methods focus on the positive words "romantic", "satisfying", "original", and "emotional"; however the AllenNLP Interpret's methods are more noisy and highlight many other words too.}
\end{figure}

\begin{figure}[H]
\begin{tabular}{m{2cm} | m{2cm}}
Shapely & \includegraphics[height=40pt]{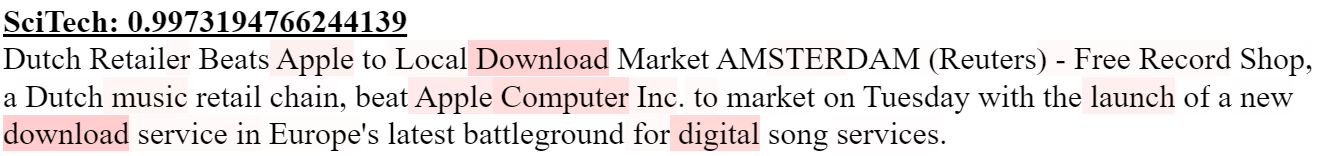} \\
Occlusion & \includegraphics[height=40pt]{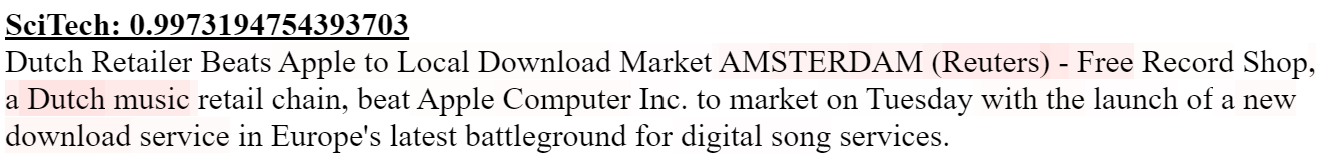} \\
Integrated & \includegraphics[height=40pt]{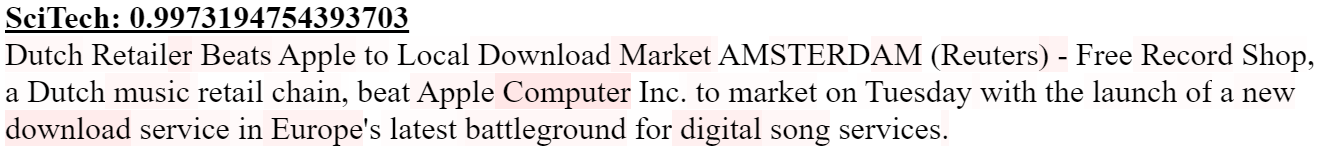} \\
Decoded Grad-CAM $l_6$ & \includegraphics[height=40pt]{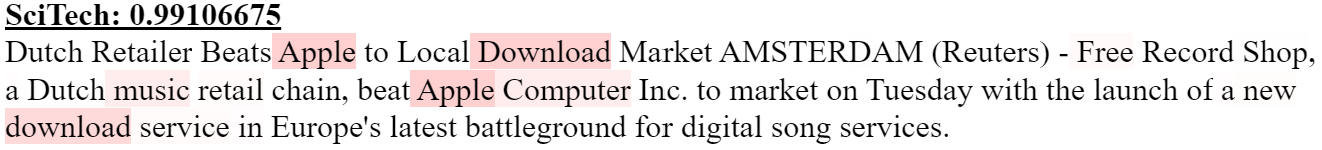} 
\end{tabular}
\caption*{AG News Example 1: Decoded Grad-CAM $l_6$ and Shapely focus on highlighting "Apple" (a tech company) along with technology terms like "Download" and "Computer". Occlusion focuses on terms related to the Netherlands such as "AMSTERDAM" and "Dutch", which do not have an obvious connection to technology. Integrated lightly highlights a large number of words, but some are technology related ones.}
\end{figure}

\begin{figure}[H]
\begin{tabular}{m{2cm} | m{2cm}}
Shapely & \includegraphics[height=40pt]{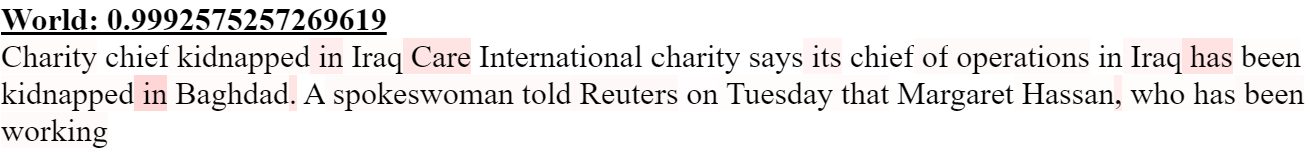} \\
Occlusion & \includegraphics[height=40pt]{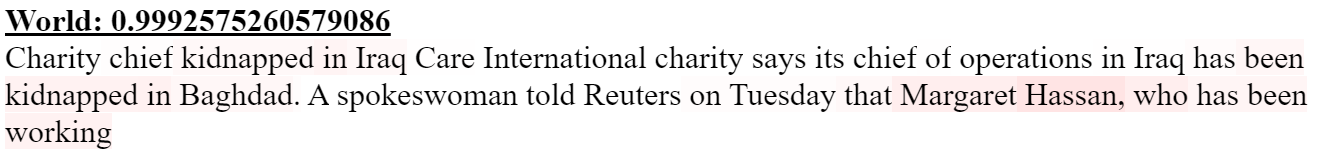} \\
Integrated & \includegraphics[height=40pt]{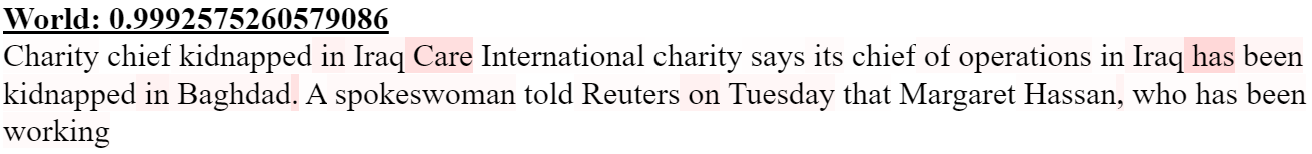} \\
Decoded Grad-CAM $l_6$ & \includegraphics[height=40pt]{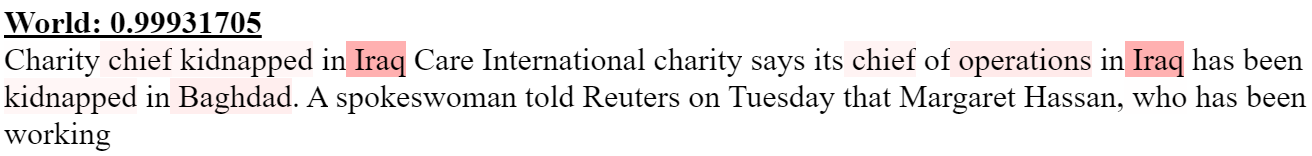} 
\end{tabular}
\caption*{AG News Example 2: Decoded Grad-CAM $l_6$ highlights the words "chief" and "kidnapped" along with terms related to the Middle East region ("Iraq" and "Baghdad"). Occlusion lightly highlights the phrases "kidnapped in Iraq" and "kidnapped in Baghdad", which also are meaningful. Shapely and Integrated have less clear explanations with focus on the words "in", "Care", and the punctuation.}
\end{figure}

\begin{figure}[H]
\begin{tabular}{m{2cm} | m{2cm}}
Shapely & \includegraphics[height=40pt]{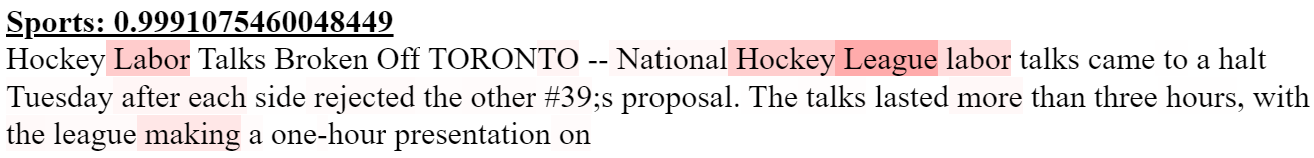} \\
Occlusion & \includegraphics[height=40pt]{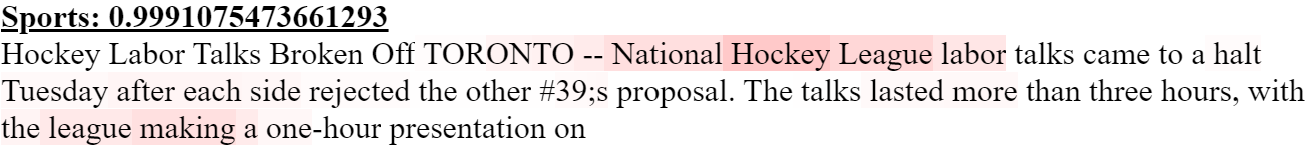} \\
Integrated & \includegraphics[height=40pt]{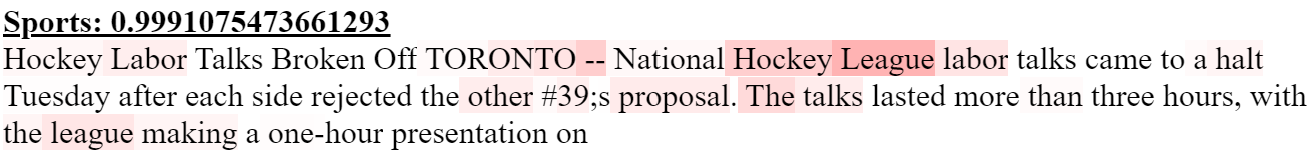} \\
Decoded Grad-CAM $l_6$ & \includegraphics[height=40pt]{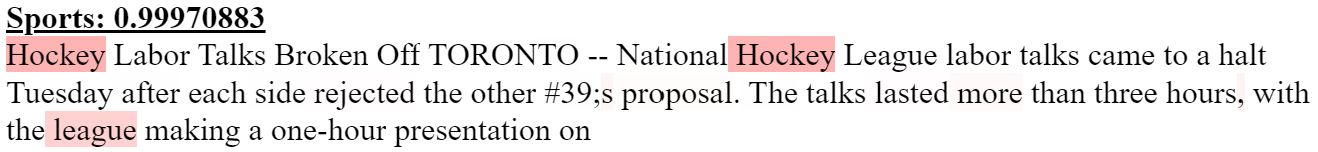} 
\end{tabular}
\caption*{AG News Example 3: Decoded Grad-CAM $l_6$ heavily highlights the word "Hockey". The other methods also have some focus on hockey terms such as the phrase "National Hockey League labor", but are noisy and also highlight many unrelated terms such as "The talks" and "after each".}
\end{figure}

\begin{figure}[H]
\begin{tabular}{m{2cm} | m{2cm}}
Shapely & \includegraphics[height=40pt]{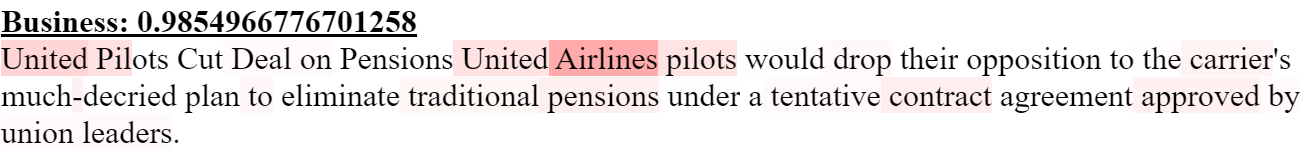} \\
Occlusion & \includegraphics[height=40pt]{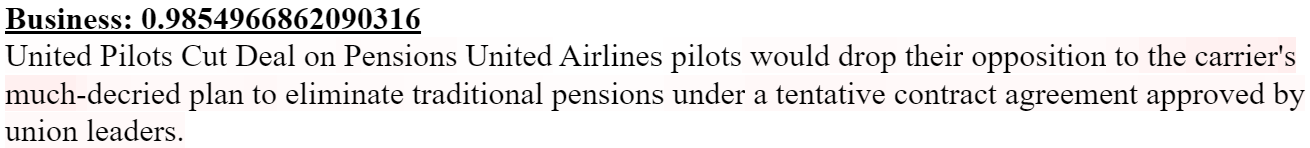} \\
Integrated & \includegraphics[height=40pt]{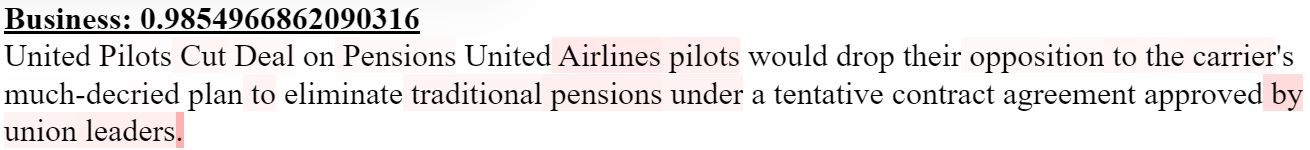} \\
Decoded Grad-CAM $l_6$ & \includegraphics[height=40pt]{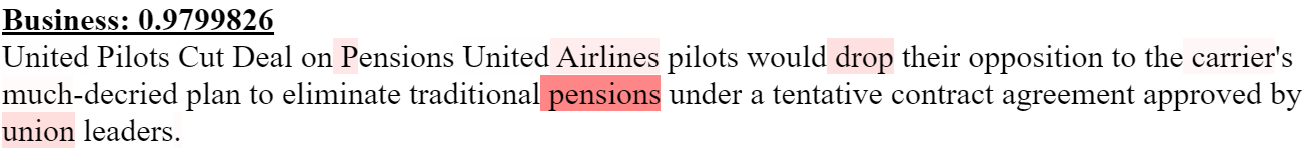} 
\end{tabular}
\caption*{AG News Example 4: Decoded Grad-CAM $l_6$ heavily highlights the word "pensions" with some additional focus on "union"; however, it also does highlight some less clear terms such as "carrier", "drop", and "Airlines". Shapely and Integrated also highlight key business terms such as "traditional pensions", "contract", and "union leaders"; although Shapely also puts a lot of emphasis on "United" and Integrated on "Airlines pilots". Occlusion lightly highlights everything and does not have an clear explanations. }
\end{figure}

\end{document}